\documentclass[twoside]{article}

\usepackage[preprint]{aistats2027}
\usepackage[round]{natbib}

\usepackage{amsmath,amsfonts,bm}

\def\1{\bm{1}}

\DeclareMathAlphabet{\mathsfit}{\encodingdefault}{\sfdefault}{m}{sl}
\SetMathAlphabet{\mathsfit}{bold}{\encodingdefault}{\sfdefault}{bx}{n}

\newcommand{\x}{{\bf x}}

\newcommand{\I}{{\bf I}}

\newcommand{\N}{\mathcal{N}}  

\newcommand{\Q}{{\bf Q}}

\newcommand{\Scal}{{\mathcal{S}}}

\newcommand{\X}{{\bf X}}

\providecommand{\1}{{\bf 1}}

\newcommand{\ben}{\begin{enumerate}}
\newcommand{\een}{\end{enumerate}}

\newcommand{\eg}{{{\textit{e.g.},}}\xspace}

\usepackage{dsfont}
\usepackage{amssymb, amsmath,amsthm}
\usepackage{amsfonts}

\usepackage{algorithm}
\usepackage{algorithmic}
\usepackage{paralist}
\usepackage{multirow}
\usepackage{booktabs}

\usepackage{times}
\usepackage{wrapfig}

\usepackage{url,enumerate}
\usepackage{color,xcolor}
\usepackage{makeidx}  
\usepackage{amsmath,amssymb}
\usepackage{mathtools}
\usepackage[small, compact]{titlesec}
\usepackage{xspace}
\usepackage{epstopdf}
\usepackage{mathrsfs}
\usepackage{times}
\usepackage{enumerate}
\usepackage{color}
\usepackage{graphicx,epsfig}
\usepackage{bm}
\usepackage{bbm}
\usepackage{upgreek}
\usepackage{multirow}
\usepackage[normalem]{ulem}
\usepackage{cancel}
\usepackage{empheq}
\usepackage{pifont}
\usepackage{subcaption}
\usepackage{booktabs, array, seqsplit}

\usepackage[capitalize,noabbrev]{cleveref}

\theoremstyle{plain}

\theoremstyle{definition}

\theoremstyle{remark}

\newcommand{\cmt}[1]{}

\newcommand{\ours}{{MDSR}\xspace}

\begin{document}

%

%

\twocolumn[

\aistatstitle{Mask-Based Deep Symbolic Regression}

\aistatsauthor{ Zachary Bastiani \And Robert M. Kirby \And Jacob Hochhalter \And  Shandian Zhe}

\aistatsaddress{University of Utah \And University of Utah \And University of Utah \And University of Utah} ]

\begin{abstract}

Symbolic regression (SR) has changed considerably in recent years, with much of the progress driven by pretrained foundation models, agentic systems, evolutionary search, and hybrid methods that combine multiple approaches. In contrast, comparatively less attention has been devoted to improving the core learning and generation mechanisms of specialized SR frameworks such as deep symbolic regression (DSR). Motivated by advances in sampling procedures for natural language processing, we revisit the DSR framework and propose Mask-based Deep Symbolic Regression (\ours). \ours introduces a masked sampling procedure together with a multi-step REINFORCE policy, substantially improving DSR and demonstrating strong SR performance without relying on pretraining, evolutionary search, or other hybrid components. \ours is competitive with leading SR methods on the SRBenchmark and produces parsimonious expressions on real scientific problems, making it well suited to challenging, data-sparse settings where interpretability is essential.

\end{abstract}

\section{Introduction} \label{sec:introduction}

Symbolic regression (SR) seeks to discover compact mathematical expressions that explain observed relationships from data. Unlike conventional predictive models with fixed architectures, SR searches over both the structure and parameters of expressions, making interpretability a central objective. This is particularly valuable in scientific applications, where a concise equation can reveal underlying relationships rather than merely provide accurate predictions. Deep Symbolic Regression (DSR)~\citep{petersen_deep_2019} provides a simple task-specific framework for this problem: an autoregressive neural policy generates candidate expressions and is optimized using risk-seeking reinforcement learning. DSR has since become an important baseline for learning-based SR, with subsequent work improving objectives such as the tradeoff between accuracy and expression complexity~\citep{bastiani2026}.

Recent progress in SR has increasingly relied on pretrained foundation models~\citep{kamienny_end--end_2022,valipour_symbolicgpt_2021} and hybrid systems that combine neural generation with genetic programming (GP), Monte Carlo Tree Search (MCTS), or other search procedures~\citep{browne2012survey}. While powerful, these approaches can require substantial offline pretraining and often retain expensive search or refinement procedures at deployment. This motivates a complementary question: \emph{how far can a task-specific SR framework be improved by redesigning its core learning and generation mechanisms, without relying on pretraining or hybrid search?} We revisit DSR from this perspective.

Our key observation is that effective symbolic search requires balancing \emph{exploration} of structurally diverse expressions with \emph{exploitation} of promising candidates. Existing neural SR samplers tend to favor one side of this tradeoff. Autoregressive generation provides strong sequential conditioning and effective exploitation, but commits early to previously generated tokens and can limit structural exploration. Diffusion-based approaches~\citep{xiang2026modsr,meidani2024snip}, in contrast, enable iterative regeneration and broader exploration, but generate tokens from more weakly conditioned intermediate states. Motivated by masked generation in NLP~\citep{makspredict2019,nie2025largelanguagediffusionmodels}, we introduce a mask-based sampler that iteratively regenerates subsets of tokens while conditioning on the remaining expression, combining strong contextual conditioning with flexible structural revision.

We further reconsider policy optimization in DSR. Multi-step trust-region methods such as PPO~\citep{schulman2017ppo} and GRPO~\citep{shao2024deepseekmath} stabilize repeated policy updates through clipping or related constraints. For risk-seeking symbolic search, however, we find that such constraints can overly concentrate the policy around high-reward samples, reducing the diversity needed for effective exploration. We instead introduce multi-step REINFORCE, which retains reward-scaled policy gradients while performing multiple updates per sampled batch. Together with mask-based sampling, this yields Mask-based Deep Symbolic Regression (\ours), which substantially strengthens task-specific SR without pretraining, evolutionary search, or other hybrid components.

Our main contributions are:
\begin{itemize}
	\item \textbf{Mask-based sampling.} We introduce an iterative random-mask expression sampler that combines the strong conditioning of autoregressive generation with iterative regeneration, improving the exploration--exploitation tradeoff over autoregressive, Discrete Denoising Diffusion Probabilistic Models (D3PM), and masked-diffusion samplers.
	
	\item \textbf{Multi-step risk-seeking REINFORCE.} We develop a multi-step, token-wise risk-seeking REINFORCE procedure that promotes exploration while preserving effective exploitation. In contrast, trust-region methods such as PPO and GRPO reduce diversity and lead to lower symbolic recovery in our experiments.
	
	\item \textbf{Strong empirical performance.} On SRBench, \ours substantially outperforms existing DSR-based methods and is competitive with leading SR approaches. On real-world problems in geology and ecology, it discovers highly parsimonious expressions that consistently lie on the Pareto frontier of predictive accuracy and expression complexity.
\end{itemize}

\section{Background} \label{sect:bk}

\noindent\textbf{Deep Symbolic Regression (DSR).} Given a measurement dataset, DSR trains a recurrent neural network (RNN) to generate expressions that describe the underlying data \citep{petersen_deep_2019}. The RNN predicts each token in the preorder traversal of the expression tree in an autoregressive manner. Training is performed via reinforcement learning, where the reward is based on the normalized root mean squared error (NRMSE) of the data fit:
$\text{NRMSE} = \frac{1}{\sigma_y}\sqrt{\frac{1}{n}\sum_{i=1}^{n}(y_i - \tau(\x_i))^2}$,
where $\tau$ denotes the expression and $\sigma_y$ is the standard deviation of the outputs in the dataset. Since the goal is to identify the best expressions rather than the best average expression, DSR employs a risk-seeking policy gradient in which only the top $\alpha\%$ of sampled expressions are used to update the model at each iteration (an entropy bonus is also added to encourage exploration):
\begin{align}
	I_{\text{fit}}(\tau) &= \frac{1}{1 + \text{NRMSE}(\tau, \x, y)}, \label{eq:dsr-reward} \\
	R(\tau) &= \text{ReLU}(I_{\text{fit}}(\tau) - I_{\text{fit}}(\tau_\alpha)) \\
	\nabla J_{\text{risk}}(\theta ; \alpha) &= \frac{100}{B\alpha} \sum\nolimits_{i=1}^{B} R(\tau^{(i)}) \nabla_{\theta} \log p_{\theta}(\tau^{(i)}), \label{eq:risk_seeking_gradient}
\end{align}

where $B$ is the number of expressions sampled at each epoch, $\tau_\alpha$ is the expression at the $(1-\alpha/100)$ quantile of the batch rewards (the lowest-reward expression among the top $\alpha\%$), and $\theta$ denotes the model's parameters.

Complexity-Aware Deep Symbolic Regression (CADSR) introduces several modifications to DSR, but we focus on four key changes: (i) a transformer-based policy operating in the frequency domain, (ii) a risk-seeking Group Relative Policy Optimization (GRPO) update, (iii) a rank-based Bayesian Information Criterion (BIC) reward that trades off data fit against expression complexity, and (iv) a memory replay buffer of high-performing expressions \citep{bastiani2026}. (i) CADSR leverages a transformer architecture while projecting the inputs to the frequency domain, with the primary benefit over DSR being that the transformer attends over all inputs rather than relying on a memory tensor to handle expression-tree information. (ii) CADSR leverages GRPO to improve training by taking multiple optimization steps per sample cycle while relying on token-wise clipping and a Kullback--Leibler (KL) divergence penalty to prevent erroneous updates. (iii) CADSR leverages a rank-based BIC reward to remain unbiased and balance parsimony of discovered expressions. (iv) CADSR uses a memory replay buffer of size $\alpha B / 100$ that is used in all optimization steps to provide consistent guidance towards top-performing expressions.

\begin{align}
    \text{BIC}(\tau) &= k \log S - 2 \log p(\tau), \\
	\gamma(\tau) &= \left|\left\{\tau' \in \mathcal{S}_\alpha : \text{BIC}(\tau') > \text{BIC}(\tau)\right\}\right|\\
    R(\tau) &= \lambda \, \text{ReLU}\!\left(1 - \frac{\gamma(\tau)}{\alpha B / 100}\right) \label{eq:bic_reward} 
\end{align}

\noindent\textbf{Token Generation Methods.} Natural language processing (NLP) represent text as a sequence of tokens, similar to how DSR treats expression. NLP has three common families for how the sequence can be generated: autoregressive sampling, masked sampling, and discrete diffusion. Masked sampling can be view as the center of these families as it has overlap with both discrete diffusion and autoregressive sampling.

Autoregressive sampling is the most common approach. Each token is sampled in left-to-right order, with each position conditioned on all previous positions. Attention-based models can be trained efficiently for this process with a causal attention mask, which allows all positions of a sequence to be trained in parallel (teacher forcing).

Discrete Denoising Diffusion Probabilistic Models (D3PM) learn to iteratively denoise a sampled sequence to align with the target distribution \citep{austin2023d3pm}.
D3PM defines $X_0$ as the uncorrupted sample and $\bar{Q}_t$ as a discrete transition matrix. The forward process can be written as $X_t \sim x_t = \bar{Q}_t X_0$, where $X_t$ is sampled from the corrupted distribution $x_t$. The learned reverse process is written as $X_0 \sim p_{\theta}(x_0 | X_t, t)$, where the model directly predicts $x_0$ at each timestep. The closed form for the reverse and forward processes of D3PM enables $p_{\theta}(x_0|X_t, t) \forall t$, rather than incrementally predicting $p_{\theta}(x_{t-1} | X_t, t)$, as the forward process can be used corrupt the predicted distribution $x_0$ to obtain the sample $X_{t-1}$.

Recently, the absorbing-state (masking) variant, known as masked diffusion, has been shown to be competitive for language modeling \citep{nie2025largelanguagediffusionmodels}.
In masked diffusion, the transition matrix is a state absorbing matrix: each token is independently replaced by a mask token with a probability that increases with diffusion time, and the reverse process unmasks tokens, possibly several per step. This follows a similar logic to D3PM, where at each timestep the model predicts $x_0$ and relies on a sampling and re-masking step (forward process) to enforce small steps forward in time. A detailed description of the implementation of all three methods is provided in Appendix \ref{sec:sampling_ablations_appendix}.

\section{Method}\label{sec:methodology}

\begin{figure}
	\includegraphics[width=0.48\textwidth]{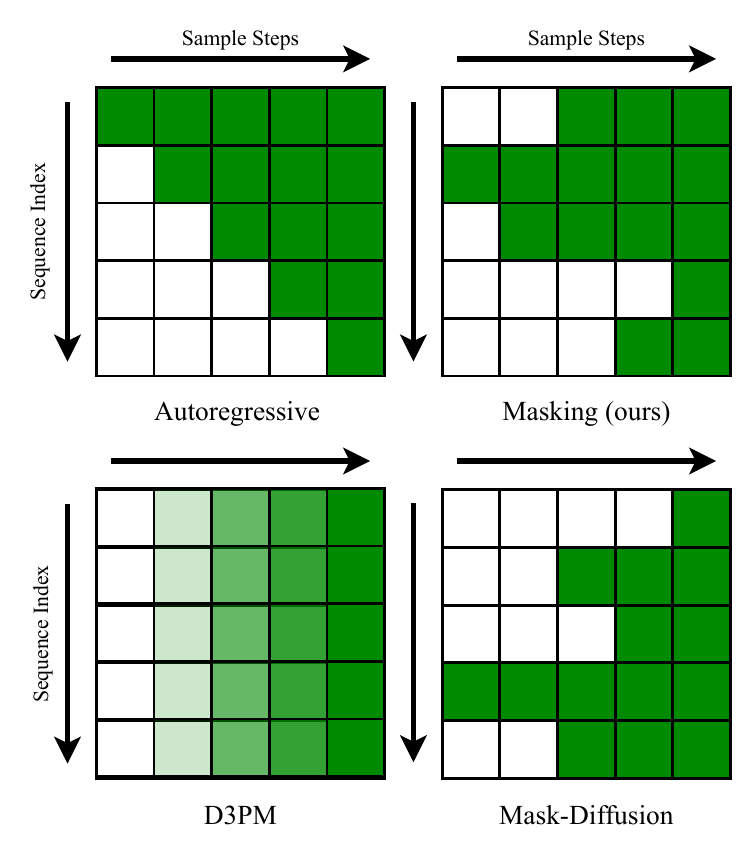}
	\caption{Visual representation of sampling methods where colored cells represent a sampled index, while white cells denote an unsampled or masked index.}
	\label{fig:sampling_procedures}
\end{figure}

\subsection{Mask-based Sampling}

In risk-seeking reinforcement learning, the sampling procedure largely determines the balance between exploration and exploitation. Our ablations (Section~5.2) show that the choice of sampler substantially changes the diversity of generated expressions, from autoregressive sampling at the low end to Mask-Diffusion at the high end. We hypothesize that this spectrum reflects how strongly each sampler conditions and commits tokens. Autoregressive sampling favors exploitation: each token conditions on all previous tokens and, once generated, is fixed, keeping samples close to trajectories the model already considers promising. D3PM shifts toward exploration because all positions remain revisable over multiple denoising steps, allowing samples to move away from early decisions; this effect becomes stronger with more timesteps. Mask-Diffusion pushes exploration further for two reasons. First, tokens are committed in a random order, so a high-entropy token may be fixed before the surrounding context has sufficiently narrowed its prediction. Second, multiple tokens are sampled simultaneously from independent per-position distributions, allowing their combination to depart from promising model trajectories. We therefore propose unmasking exactly one token per step in a random order, preserving flexible exploration while recovering the strong conditioning that enables effective exploitation in autoregressive sampling.

We represent a symbolic expression as a token matrix $\X \in \{0,1\}^{M \times d}$, where each row is the one-hot encoding of a token, $d$ is the vocabulary size, and $M$ is the maximum number of tokens. Unsampled positions are represented by all-zero rows, which act as the mask. We initialize $\X_{M}$ as the zero matrix and let $\theta$ denote the weights of our decoder-only transformer-based network.

To sample an expression, we take an approach similar to existing masking methods: at each step the model predicts a full expression $\X_0$, and we then mask a portion of it. We divide sampling into $M$ steps and require that exactly one new position is unmasked at each step, so that $t \in \{1, \dots, M\}$. At each step, the model $p_\theta$ outputs a categorical distribution for every position in the sequence for which we sample a sequence of tokens, as shown in Equation~\ref{eq:mask_sampling_eq}. We then apply the remasking step in Equation~\ref{eq:masking_process}, which keeps a single newly sampled token and re-masks the rest, so that tokens unmasked at earlier steps remain fixed throughout sampling. This makes our masking approach match the standard autoregressive approach in requiring $M$ sequential steps. We can write this similarly to the masked diffusion process, as we define $\bar{Q}_t = Q_{t} + Q_{t+1} + \cdots + Q_{M}$, where $Q_t$ unmasks one token and $Q_i \neq Q_j \; \forall i \neq j$. The proposed masking process is a more constrained version of the masked diffusion process and less constrained than autoregressive sampling, enabling more diverse sampling with strong exploitation. 

\begin{align}
	\X_0 &\sim p_{\theta}(\x_0 | \X_{t}, t) \label{eq:mask_sampling_eq}\\
	\X_{t-1} &= Q_{t-1} \X_0 + \bar{Q}_t X_t \label{eq:masking_process}
\end{align}

where $\X_{t}$ contains the tokens sampled at positions $\bar{Q}_{t}$ and zeros everywhere else. One key detail to note is that at each step the model is predicting $\X_0$, thus enabling the log probability $\log p_{\theta}(\X_0 | \X_{t}, t) $. During training, we sample a random timestep $t \in [1, M]$ and randomly mask out $t$ tokens.

This procedure differs from the alternatives in how positions are selected. Mask-Predict-style methods use the model's confidence to choose which positions to fill, which is more exploitative and may reduce sample diversity. Masked diffusion can fill several positions per step independently, even if the number of timesteps is set to the sequence length, which leaves those tokens less correlated than if each were conditioned on the others. Our random-order, one-token-per-step scheme conditions every token on all previously sampled tokens while still visiting positions in varied orders. We hypothesize that this balances the diversity of masked diffusion against the exploitation of autoregressive sampling, and we test this hypothesis empirically in Section~\ref{sec:ablations} and provide additional support in Appendix~\ref{sec:sampling_ablations_appendix}.

Because tokens are sampled in arbitrary order, the partially filled expression is often an incomplete or disconnected tree, so we use the row index of the matrix as the token's position. We align the expression tree ordering with CADSR and use the breadth-first search ordering. We encode both the index of the token and the current timestep into a 2D positional encoding in Equation~\ref{eq:positional_encoding}.
We follow this convention from masked diffusion; however, because our method samples one index at a time, we do not expect it to have a large impact on performance.

\begin{align}
	\text{PE}(l, t)_{2i} &= \sin(\frac{l}{10000^{(4i/D)}}), \\
	\text{PE}(l, t)_{2i+1} &= \cos(\frac{l}{10000^{(4i/D)}}), \\
    \text{PE}(l, t)_{D+2j} &= \sin(\frac{t}{10000^{(4j/D)}}), \\
	\text{PE}(l, t)_{D+2j+1} &= \cos(\frac{t}{10000^{(4j/D)}}), \label{eq:positional_encoding} 
\end{align}

The network is a transformer whose self-attention is not restricted by position: each position may attend to all other positions, including unsampled ones (mask rows carry no token information). Thus, we leverage a bidirectional attention mask to enable attention across the full sequence. 
Expression-validity constraints that DSR enforces during preorder sampling (e.g., arity constraints) cannot be applied directly in random order; we therefore apply no constraints during the sampling process. Once sampling has completed, if a tree would remain incomplete due to sampling the predicted token, we randomly sample a replacement token from the leaf-node token library ($x_0, x_1, C, \dots$).

\subsection{Multi-Step REINFORCE}

In reinforcement learning, the model is a policy that selects actions to maximize expected reward. Policy-gradient methods aim to increase the probability of samples with positive advantages while reducing the probability of samples with negative advantages. The advantage of a sampled expression $\tau^{(i)}$ measures how much better it is than a baseline; we use the rank-based BIC reward of CADSR from Equation~\ref{eq:bic_reward} as this advantage.

The original REINFORCE policy optimization uses each batch of on-policy samples for a single gradient step. To reuse samples more efficiently, Trust Region Policy Optimization (TRPO) \citep{schulman2017trpo} maximizes a surrogate objective weighted by the probability ratio $\rho_i= p_{\theta}(\tau^{(i)}) / p_{\theta_{\text{old}}}(\tau^{(i)})$ between the current and previous policies, subject to a KL-divergence constraint that keeps the new policy close to the old one. Proximal Policy Optimization (PPO) \citep{schulman2017ppo} replaces the hard constraint with a simpler clipped objective, in which samples whose ratio leaves $[1-\epsilon,1+\epsilon]$ in the direction favored by the advantage contribute no gradient. GRPO \citep{shao2024deepseekmath} is a variant of PPO that removes the learned value function and computes advantages by normalizing rewards within a group of sampled outputs.

Compared with REINFORCE, PPO-style methods differ in two ways: (i) the clipping/trust region that limits how far the policy can move on a fixed batch, and (ii) the scaling of each sample's gradient by the ratio $\rho_i$, since $\nabla_\theta \rho_i = \rho_i \nabla_\theta \log p_{\theta}(\tau^{(i)})$.

We propose token-wise multi-step REINFORCE: we take $C$ gradient steps on the same set of top-$\alpha\%$ samples using the plain REINFORCE gradient, with no ratio or clipping,
\begin{align}
	\nabla_\theta J_{\text{MS}} = -\frac{100}{B\alpha}\sum_{i=1}^{B} R(\tau^{(i)}) \nabla_\theta \log p_{\theta}(x_0|X_t, t). \label{eq:ms-reinforce}
\end{align}

This is riskier than PPO. After the first step, the samples are no longer drawn from the current policy, so the estimator is biased, and nothing prevents large or erroneous updates or overfitting to the batch. We accept this trade-off because our objective is risk-seeking: we care about finding the best expressions, not about an unbiased estimate of the expected reward, and larger updates can move the policy toward high-quality expressions faster. We evaluate this choice against PPO- and GRPO-style updates in Section~\ref{sec:ablations}.

\begingroup
\setlength{\textfloatsep}{0pt}
\begin{algorithm}[!t]
	\caption{Mask-based Deep Symbolic Regression}
	\label{alg:ddsr}
	\textbf{input} Learning rate $\gamma$; risk factor $\alpha$; expression batch size $B$; gradient steps per epoch $C$; number of epochs $N$; entropy scalar $\beta$ \\
	\textbf{output} The best expression $\tau^*$
	\begin{algorithmic}[1]
		\STATE Initialize transformer $\phi$ with parameters $\theta$
		\STATE $\Scal_\alpha \gets \{\}$, \; $\tau^* \gets \text{None}$
		\FOR{$i = 0$ to $N-1$}
		\STATE Sample $B$ expressions (each with a random order $\sigma$) using the current model, and obtain the top $\alpha\%$ expressions $\Scal^i_\alpha$
		\STATE $\Scal_\alpha \gets \Scal_\alpha \cup \Scal^i_{\alpha}$; update $\tau^*$ with the best expression seen so far
		\FOR{$j = 1$ to $C$}
		\STATE Sample $t \sim [1, M]$
		\STATE Compute $\nabla_\theta J_{\text{MS}}$ from~\eqref{eq:ms-reinforce} on $\tau_{\alpha}$ with $t$
		\STATE $\theta \gets \theta + \gamma (\nabla_{\theta} J_{\text{MS}} + \beta\cdot \text{Entropy-Gradient})$
		\ENDFOR
		\STATE Remove the bottom half of expressions from $\Scal_\alpha$ according to their rewards
		\ENDFOR
		\STATE \textbf{return} $\tau^*$
	\end{algorithmic}
\end{algorithm}
\endgroup
\section{Related Work}
The Deep Symbolic Regression (DSR) framework~\citep{petersen_deep_2019} pioneered the use of reinforcement learning to train RNN-based expression generators from measurement data. Building on this framework, recent extensions~\citep{tenachi2023deep, jiang2024vertical} have either enforced physics-unit constraints as domain knowledge to enhance the quality of expression generation or confined the search space through vertical discovery strategies for vector symbolic regression. Complexity-Aware Deep Symbolic Regression (CADSR)~\citep{bastiani2026} was introduced as an improvement to DSR, leveraging an improved policy architecture and a parsimonious, rank-based reward function based on the Bayesian information criterion (BIC). Our proposed multi-step REINFORCE policy aligns with CADSR's idea to leverage GRPO, but 

Another line of work has shifted toward building foundation models that map numerical measurements directly to symbolic expressions~\citep{biggio2021neural, kamienny_end--end_2022, valipour_symbolicgpt_2021, vastl_symformer_2022}. These models typically adopt an encoder-decoder transformer architecture, where encoder layers extract structural patterns from the input data and decoder layers synthesize symbolic outputs. Recently, diffusion-based foundation models~\citep{han2025discoveringmathematicalequationsdiffusion,tymkow2025symbolicdiffusiondeeplearningbased,xiang2026modsr} have shown performance competitive with autoregressive approaches. Although all of these approaches are promising, their training is costly and acutely sensitive to data preparation pipelines, often requiring massive synthetic datasets. More critically, without a data-specific search mechanism, such models often struggle to generalize---especially when faced with out-of-distribution (OOD) measurement datasets~\citep{kamienny2023deepgenerativesymbolicregression}. Recently, Multi-Modal Symbolic Regression (MMSR) was published as a new foundation model trained via contrastive learning to predict an expression skeleton before sampling operator and variable tokens \citep{Li_2025_MMSR}.

To overcome these limitations, a new wave of research couples pretrained foundation models with explicit search or planning mechanisms tailored to the target dataset, for instance, TPSR~\citep{shojaee_transformer-based_2023}, SNIP~\citep{meidani2024snip}, and MODSR~\citep{xiang2026modsr}. In TPSR, the pretrained model is integrated into a modified Monte Carlo Tree Search (MCTS) method~\citep{browne2012survey}, using model-guided token selection and tree expansion driven by an upper confidence bound (UCB) heuristic. MODSR and SNIP improve performance on OOD datasets by augmenting their methodology with a search-based algorithm built around their foundation model. MODSR is a diffusion-based foundation model that uses gradient information about how well an expression tree will fit the dataset to guide its diffusion process~\citep{xiang2026modsr}. SNIP leverages a contrastive foundation model with a latent-space representation of the dataset and the expression to search for an expression that aligns with the dataset. 
Beyond these model-based approaches, ensemble frameworks such as uDSR~\citep{landajuela2022a} combine multiple symbolic regression strategies (\eg GP and DSR) to boost robustness and accuracy. Meanwhile, model-free methods~\citep{sun2023symbolic, xu2024reinforcement} explore purely search-based techniques, using MCTS or ensemble strategies combining MCTS with GP, to uncover symbolic expressions without relying on heavy pretraining.

Recently, \citet{nie2025largelanguagediffusionmodels} proposed a masked discrete diffusion framework, wherein each token is given a probability of being unmasked that gradually increases during the reverse process. The chance of any token being unmasked is independent of the other tokens. In contrast, existing masking approaches such as Mask-Predict~\citep{makspredict2019} and MaskGIT~\citep{maskgit2022} leverage knowledge about each token's distribution to determine which tokens to unmask at each step. Our approach adopts a similar masking idea but differs in two key aspects: (1) at each step, we unmask only a single token to preserve structural information more effectively, and (2) we integrate the process into a reinforcement learning framework to guide training toward high-reward expressions.

\section{Experiments}

\begin{figure*}[ht]
    \centering
    \includegraphics[width=0.95\textwidth]{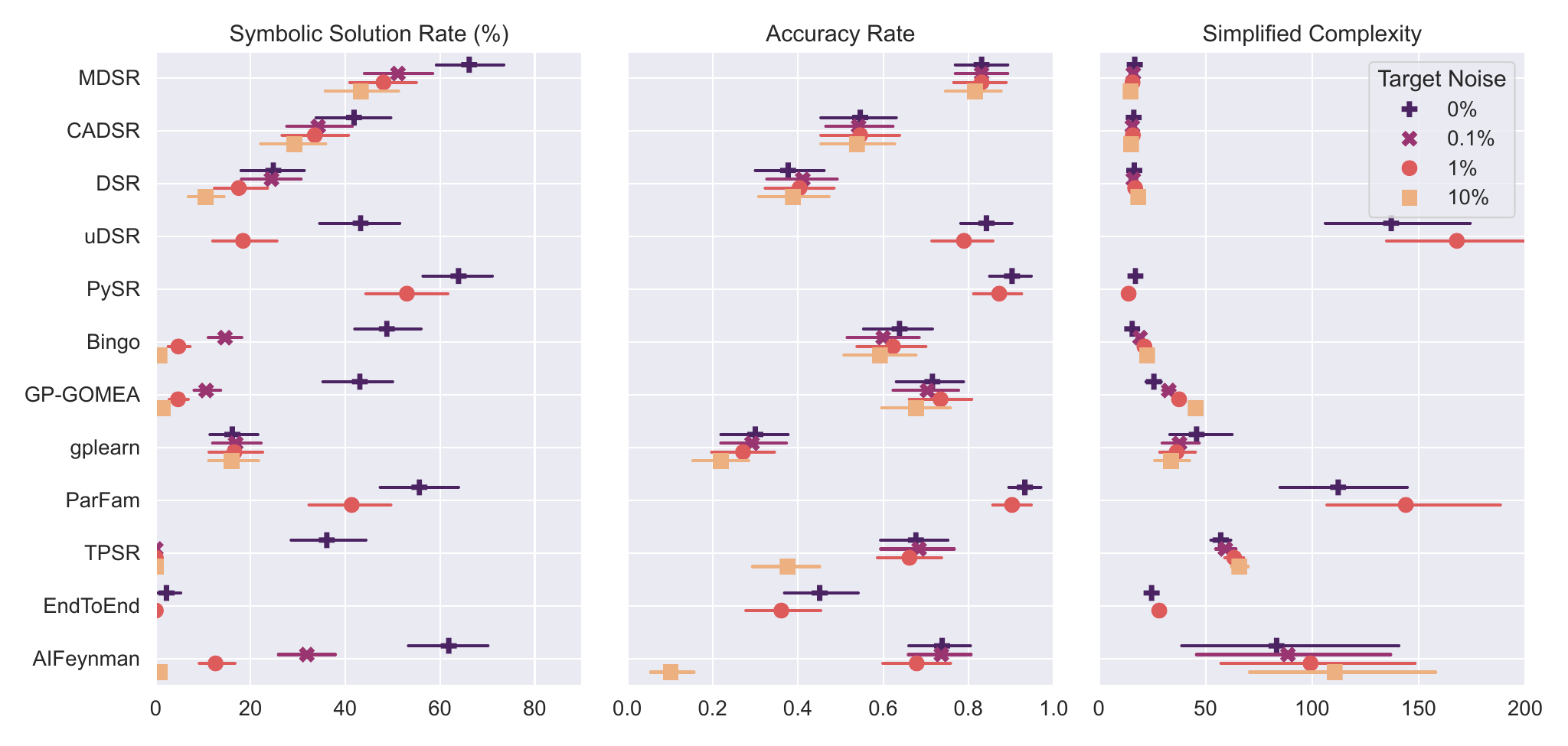}
    \caption{\ours's performance on the SRBenchmark. Methods with * are taking from ParFam's published results.}
    \label{fig:srbench_results}
\end{figure*}

We benchmarked \ours across the SRBench dataset to enable a comprehensive evaluation of its performance alongside two key ablation studies to highlight the significance of each component. Additionally, we applied \ours to two real-world applications to validate its effectiveness compared to modern symbolic regression methods.

\subsection{Performance on SRBench}

We evaluate \ours on SRBench to compare directly with CADSR and DSR. SRBench comprises two datasets: the Feynman dataset (119 problems) and the Strogatz dataset (14 problems). We ran each experiment for 500 epochs with a batch size of 1000, matching the standard budget of 500k expression evaluations specified in SRBench. We ran eight trials per problem at each of the four standard noise levels (0\%, 0.1\%, 1\%, and 10\%). To reduce GPU costs, we ran eight experiments concurrently on a single A40 GPU, which increased runtime due to resource contention. Methods are evaluated on symbolic solution rate (rate of expression recovery or $R^2 = 1.0$), accuracy rate (rate of $R^2>0.999$), and simplified complexity (number of nodes in simplified expression).

We compare against the following methods: CADSR, DSR, uDSR, PySR, Bingo, GP-GOMEA, gplearn, ParFam, TPSR, EndToEnd, AIFeynman, MODSR, MMSR, and SNIP. CADSR and DSR are the most comparable methods, as both are DSR-based reinforcement learning approaches like \ours. PySR, Bingo, GP-GOMEA, and gplearn are GP-based approaches. AIFeynman searches for hyperplanes within the dataset and fits each with a polynomial. EndToEnd, MODSR, MMSR, SNIP, and TPSR build on a pretrained foundation model, with SNIP and TPSR using a search method to improve the model's predictions. uDSR is a hybrid method that combines AIFeynman, DSR, and GP.

\ours significantly outperforms CADSR and DSR across all noise levels and categories, a substantial improvement over DSR-style methods. Compared to CADSR, \ours improves accuracy rate by roughly 35 percentage points across all noise levels, and by over 40 percentage points compared to DSR. \ours also improves symbolic recovery rate by nearly 25 percentage points at 0\% noise and by at least 14 percentage points across all noise levels compared to CADSR.

\ours is also competitive with recent methods such as ParFam and PySR in symbolic recovery rate, accuracy, and complexity. PySR is a leading GP method and ParFam is a recent EQL-based method; both achieve slightly higher accuracy than \ours. However, ParFam produces long, uninterpretable expressions, and PySR required roughly 8 hours of training. The three methods thus reach similar accuracy by different means, with each making a different trade-off among accuracy, complexity, and runtime. \ours offers low complexity and low runtime at some cost in accuracy; ParFam offers high accuracy and low runtime at the cost of complexity; PySR offers high accuracy and low complexity at the cost of runtime. These trade-offs are characteristic of DSR, EQL, and GP-based methods, respectively. Finally, \ours outperforms uDSR in symbolic solution rate, runtime, and complexity. This suggests that DSR-based methods can surpass ensemble approaches, and that embedding \ours within such ensembles could improve them further.

\begin{table*}[!h]
  \centering
  \caption{Performance comparison of different SR methods on SRBench datasets. The table shows mean $R^2$ scores (higher is better) and average number of nodes $\mathcal{C}$ (lower is better) for each method. All results are compiled from the method's published benchmark results.}
  \label{tab:foundation_model_srbench_results}
  \small
  \setlength{\tabcolsep}{4.5pt}
  \begin{tabular}{ll cc cc cc cc cc cc cc}
  \toprule
  \multirow{2}{*}{Dataset}
    & \multicolumn{2}{c}{\ours}
    & \multicolumn{2}{c}{CADSR}
    & \multicolumn{2}{c}{MODSR}
    & \multicolumn{2}{c}{MMSR}
    & \multicolumn{2}{c}{SNIP}
    & \multicolumn{2}{c}{TPSR} \\
  \cmidrule(lr){2-3} \cmidrule(lr){4-5} \cmidrule(lr){6-7}
  \cmidrule(lr){8-9} \cmidrule(lr){10-11} \cmidrule(lr){12-13} 
  & $R^2$ & $\mathcal{C}$ & $R^2$ & $\mathcal{C}$ & $R^2$ & $\mathcal{C}$
  & $R^2$ & $\mathcal{C}$ & $R^2$ & $\mathcal{C}$ & $R^2$ & $\mathcal{C}$ \\
  \midrule
  Feynman  & 0.991    & 17.7   & 0.934 & 16.5   & 0.788 & \textbf{16.4} & 0.991 & 20.8 & 0.985 & 31.6 & \textbf{0.993}  & 57.2 \\
  Strogatz & 0.994    & \textbf{13.0}  & 0.980  & 15.4   & 0.970 & 15.8 & 0.982 & 21.6 & \textbf{0.995} & 29.3 & 0.965  & 56.1 \\
  \bottomrule
  \end{tabular}

\end{table*}

We compare against foundation-model-based methods in Table \ref{tab:foundation_model_srbench_results}, using $R^2$ and complexity as the primary metrics. \ours achieves $R^2$ comparable to SNIP, MMSR, and TPSR while producing lower-complexity expressions. On the Feynman dataset, \ours produces more complex expressions than CADSR and MODSR, but with a significant gain in $R^2$. Although CADSR lagged behind modern foundation-model methods, \ours closes much of that gap, making it a competitive choice for symbolic regression.

\subsection{Ablations} \label{sec:ablations}

\begin{figure}[!ht]
  \centering
  \includegraphics[width=0.45\textwidth]{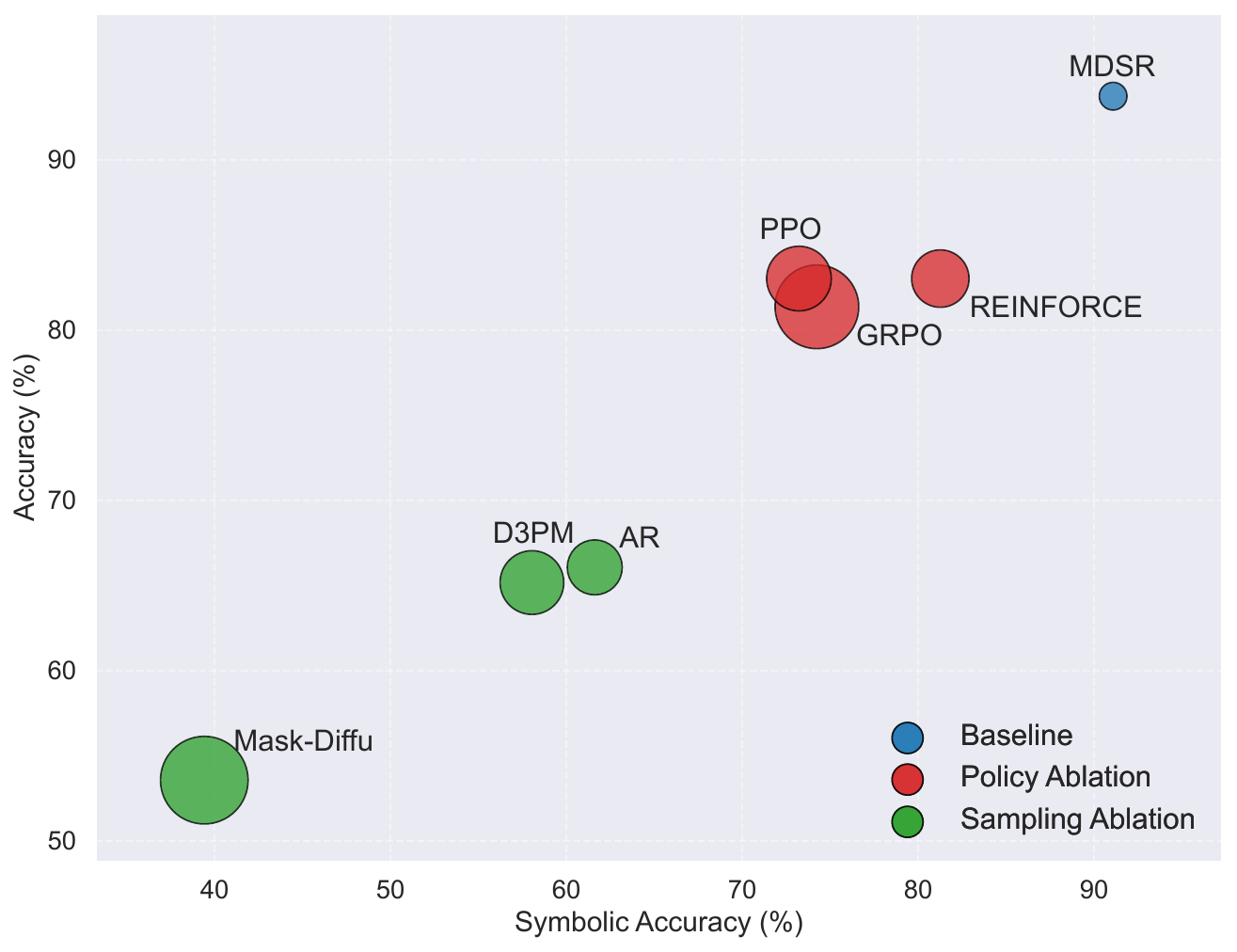}
  \caption{Ablations on the Strogatz dataset. Sampling-method ablations are shown in green and policy-optimization ablations in red; \ours (blue) serves as the baseline for both. Dot size corresponds to mean expression complexity, so smaller dots indicate more interpretable expressions.}
  \label{fig:ablations_scatterplot}
\end{figure}

To evaluate the contribution of each component of \ours, we ran six ablations on the Strogatz dataset, using eight trials per problem at 0\% noise. The ablations fall into two categories: sampling method and policy optimization. For sampling, we compared our masked sampling against standard autoregressive sampling (AR), a discrete diffusion model (D3PM), and a masked diffusion (Mask-Diffu).
For policy optimization, we compared our multi-step REINFORCE approach against a single-step REINFORCE baseline, GRPO, and PPO. Results are shown in Figure~\ref{fig:ablations_scatterplot}, and we discuss each category below.  Implementation details for each ablation are in Appendix \ref{sec:ablations_appendix}.

\begin{figure}[!b]
  \centering
  \includegraphics[width=0.45\textwidth]{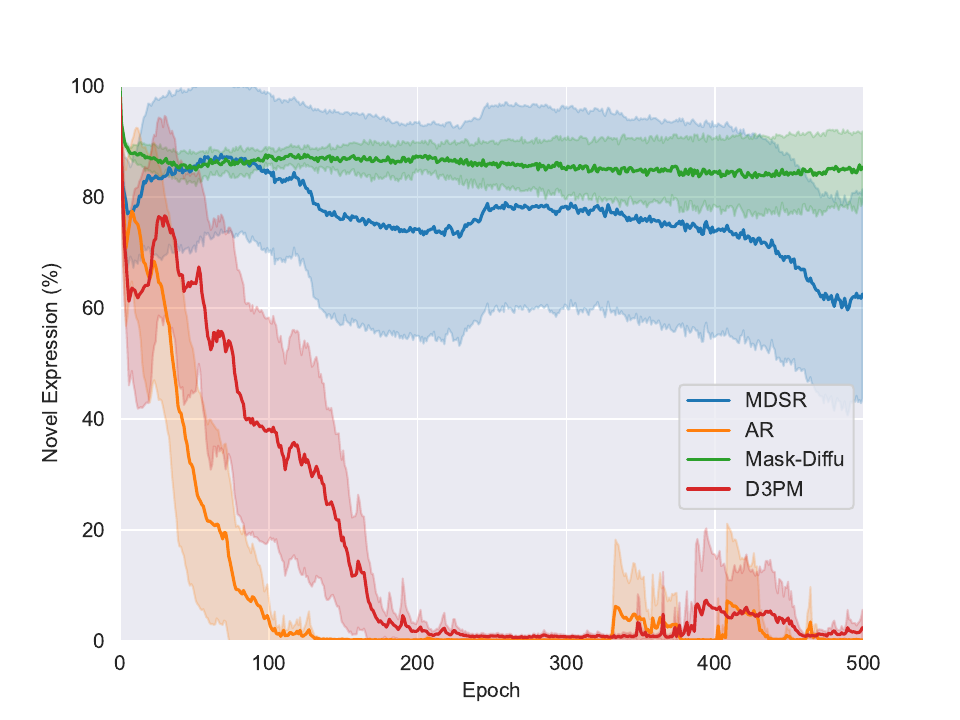}
  \caption{Comparison of the percentage of new expression found in each epoch. Shaded regions denote a 95\% confidence region.}
  \label{fig:sampling_impact_on_divesity}
\end{figure}

\textbf{Sampling Methods.} \ours outperforms all alternative sampling methods in symbolic recovery rate, accuracy rate, and expression complexity. To understand why, we also examine how each sampling strategy affects the diversity of the expressions generated. Mask-Diffu produces the most diverse expressions, but it does not sufficiently exploit promising candidates and therefore fails to find optimal solutions within the allotted compute budget. Conversely, AR and D3PM yield more consistent performance but with markedly lower diversity as seen in Figure \ref{fig:sampling_impact_on_divesity}. Notably, D3PM is more diverse than AR without achieving any improvement in performance, suggesting that its additional diversity is superfluous and potentially wasteful. In contrast, our masking approach maintains high diversity throughout training while efficiently exploiting, as reflected by its superior results. 

\textbf{Policy Optimization.} We compared our multi-step policy optimization against a single-step REINFORCE baseline, PPO, and GRPO. All three alternatives achieved an accuracy rate at least 10 percentage points lower than multi-step REINFORCE, and all were worse in complexity, accuracy rate, and symbolic recovery rate. Among them, single-step REINFORCE performed best, achieving a higher symbolic recovery rate than PPO and GRPO while maintaining complexity similar to PPO. PPO and GRPO performed comparably, differing only in a small trade-off: GRPO attained slightly higher symbolic recovery and complexity but lower accuracy than PPO. These results show that multi-step REINFORCE is the most effective method for optimizing the model as it best aligns the model with the BIC reward, seen by the reduction complexity while improving both accuracy and symbolic recovery compared to other policy optimization methods. Overall, our multi-step policy optimization contributes substantially to the performance of \ours.

\begin{figure*}[!ht]
  \centering
  \begin{subfigure}[b]{0.48\linewidth}
    \centering
    \includegraphics[width=\linewidth]{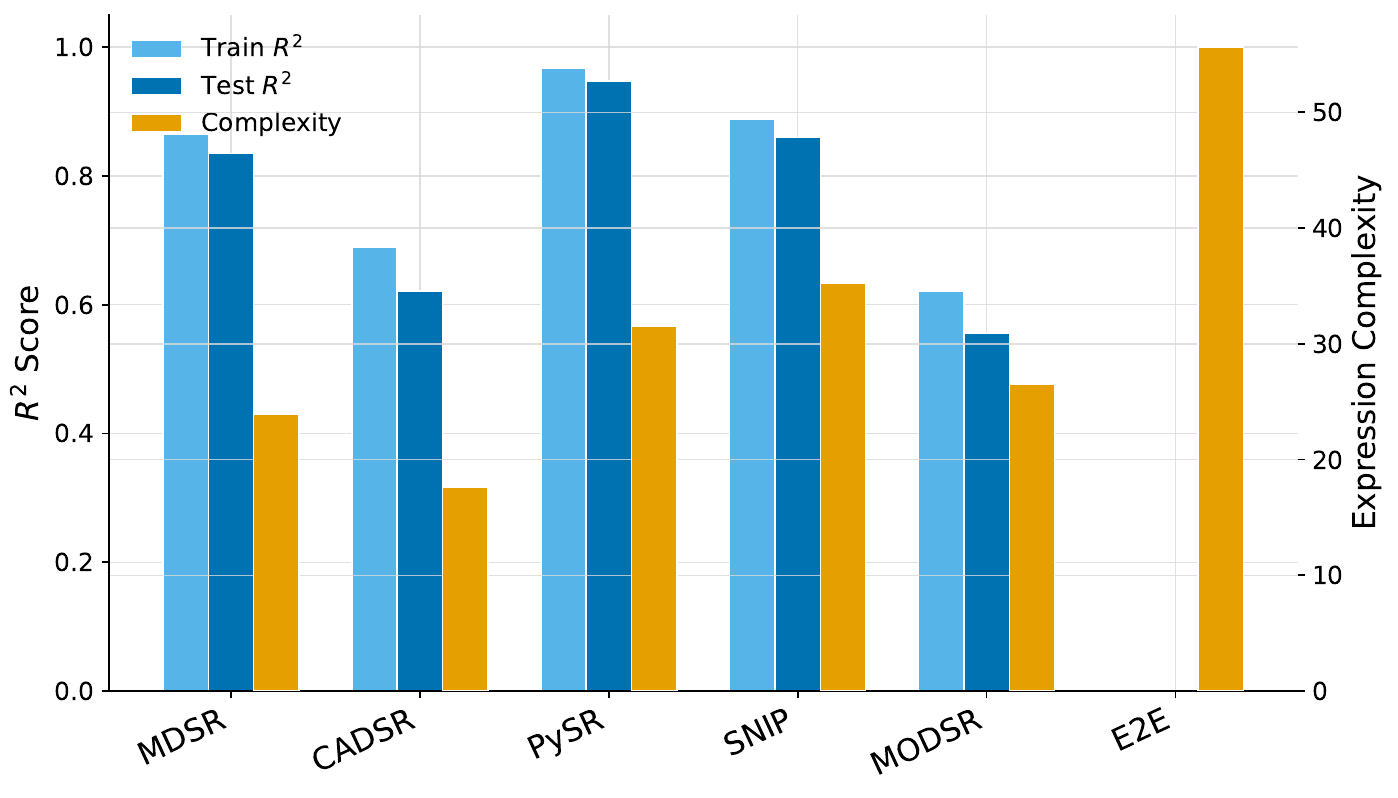}
    \caption{Mineral Retention}
    \label{fig:water_retention}
  \end{subfigure}
  \hfill
  \begin{subfigure}[b]{0.48\linewidth}
    \centering
    \includegraphics[width=\linewidth]{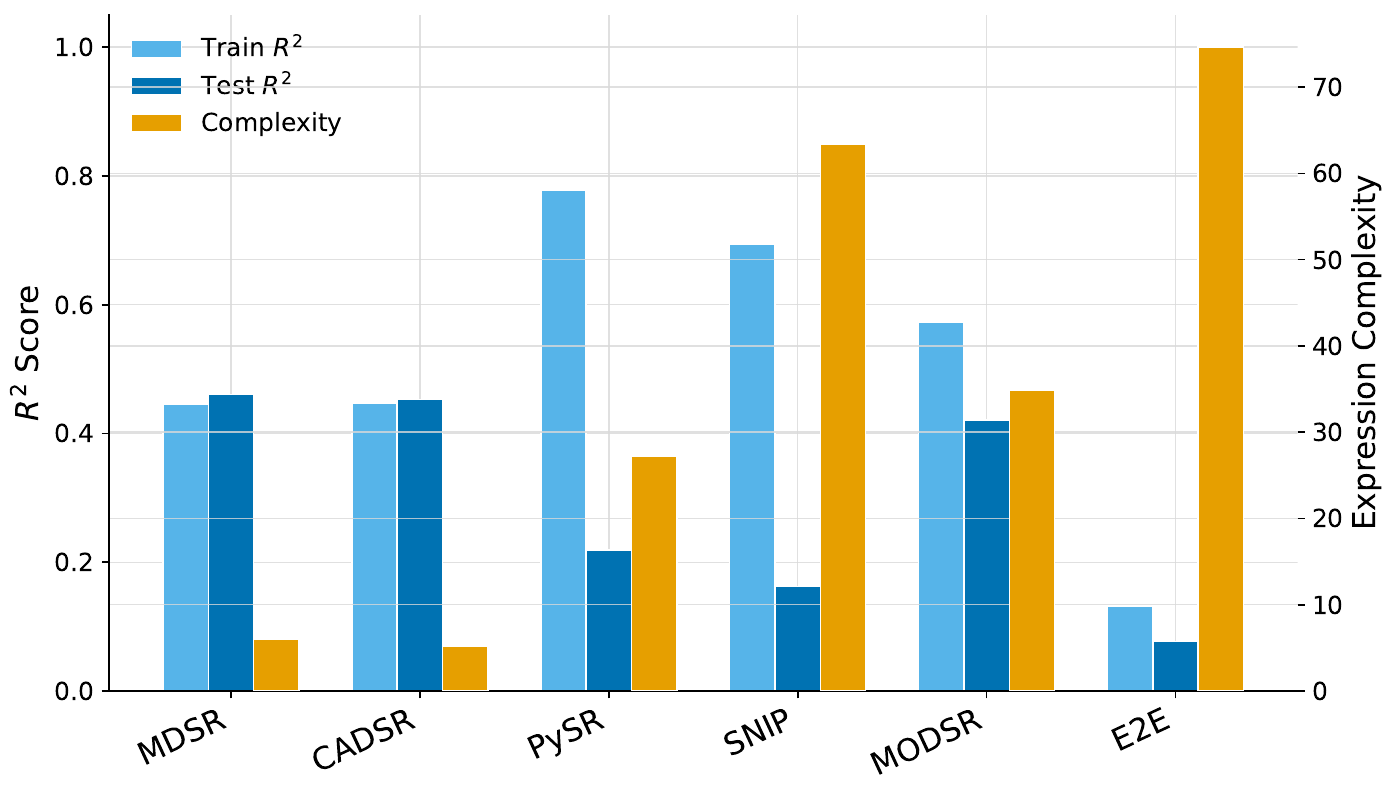}
    \caption{Genetic Expression}
    \label{fig:woodrat}
  \end{subfigure}
  \caption{Results for six symbolic regression methods across two real-world applications. Eight trials were ran for each method with access to a RTX 4090 GPU and Ryzen 9-7950X processor.}
  \label{fig:benchmark_summary}
\end{figure*}

\subsection{Mineral Retention and Genetic Expression} \label{sec:realworld_problems}

Finally, we applied \ours to two real-world problems to validate its effectiveness against modern symbolic regression methods. We compare against E2E, SNIP, MODSR, CADSR, and PySR, which together span several families of symbolic regression approaches. For every method, we used the default hyperparameters recommended by its authors; details are provided in Appendix~\ref{sec:supplymental_hyperparameter}.

\textbf{Datasets.} The first problem comes from geology. A semi-porous material can act as a filter that removes minerals from water, and the task is to discover the relationship between the filter's properties and the amount of minerals it retains \cite{water_rention_2004}. The dataset has one input feature, a cumulative variable for the amount of water added to the filter so far, and five additional features describing properties of the filter or the minerals. It contains six experiments with approximately 30 measurements each, resulting in 180 total data points. The second problem comes from ecology, where scientists measured the body mass of woodrats (log MTD mass) alongside gene sequencing to study the relationship between the two \cite{woodrats2025}. Scientists are trying to learn a relationship between genes that improve creosote resistance and the overall success of the specimen, so any relationship between these genetic factors and specimen mass is weak and noisy. The dataset contains 101 specimens, each with ten features: a global genetic measure of creosote resistance and eight gene expression strengths correlated with creosote resistance.

\textbf{Mineral Retention.} On the mineral retention problem (Figure~\ref{fig:srbench_results}), PySR finds the most accurate expression, but at relatively high complexity. \ours balances accuracy and parsimony, achieving the third-highest accuracy with nearly 8 fewer tokens than PySR. Compared to CADSR, \ours achieves higher accuracy at the cost of higher complexity. SNIP, CADSR, and \ours together form the Pareto frontier, giving practitioners a choice of trade-off between accuracy and complexity. SNIP is a potential foundational model alternative to \ours and PySR, but fall short of the Pareto fronter. MODSR achieves decent accuracy with low complexity but is outperformed by the other methods with its primary benefit being its quick runtime. E2E performs worst, as it fails to consistently produce expressions with $R^2$ above 0.

\textbf{Gene Expression.} On the gene expression problem, no method exceeds a test $R^2$ of 0.6, which reflects the difficulty of the problem. SNIP and PySR overfit substantially, with a train-to-test $R^2$ gap of about 0.5, while MODSR shows a smaller gap of about 0.15. In contrast, CADSR and \ours score higher on the test set than on the training set, and overall they produce the best expressions for this dataset. We hypothesize that this is due to their reward functions, which balance fit against dataset size, whereas SNIP and PySR apply weaker regularization. Interestingly, MODSR provides expression with comparable accuracy to \ours and CADSR, but has significantly higher complexity, resulting in diminished interpretability. The best expression found by each method, more detail descriptions of the problems, and explicit results are provided in Appendix~\ref{sec:realworld_appendix}.

\section{Conclusion} \label{sec:conclusion}

%
%

We introduced \ours, a mask-based model using multi-step REINFORCE for symbolic regression. Extensive experiments on the SRBench benchmark demonstrate that \ours consistently outperforms existing model-based reinforcement learning approaches and achieves performance comparable to state-of-the-art genetic programming (GP) and foundation model-based methods. Additionally, \ours provides parsimonious expressions for real-world problems, enabling scientists and engineers to access interpretable equations for sparse and noisy problems. Beyond this, we aim to extend our conceptual understanding to foundational models, focusing on improving their applications to real-world problems by improving parsimony. 


\newpage

\subsection*{AI use statement}

In this work, we used generative AI tools for debugging code, provide small code snippets, and setup existing methodologies. Specifically, generative AI was used to quickly download and integrate methods tested on the real-world problems to enable quick comparison to existing methods, and to review sections of the code for bugs. 
We have not used generative AI tools for generate synthetic data sets; help develop theoretical models or conceptual frameworks; formulate mathematical claims; provide critical ingredients for proving mathematical claims; assist in the writing of proofs; propose or refine hypotheses; design or provide feedback on research methodology or experiments; implement methods; assist with translation; clean and reformat datasets; support qualitative and thematic data analysis; interpret results.

We used generative AI tools for generating figures, generating tables, provide latex formatting, and help polishing manuscript. We did not use generative AI tools to draft the paper or sections of the paper, but leveraged generative AI to review and polish paragraphs within the manuscript before additional revisions by the authors.
We have reviewed all AI-assisted work to ensure it aligns with the results and authors' intended claims.
We take responsibility for the final content of this work, including text, claims, or artifacts produced with the aid of generative AI.

\bibliographystyle{apalike}
\bibliography{references}

\section*{Checklist}

\begin{enumerate}

  \item For all models and algorithms presented, check if you include:
  \begin{enumerate}
    \item A clear description of the mathematical setting, assumptions, algorithm, and/or model. [Yes]
    \item An analysis of the properties and complexity (time, space, sample size) of any algorithm. [Yes]
    \item (Optional) Anonymized source code, with specification of all dependencies, including external libraries. [Yes]
  \end{enumerate}

  \item For any theoretical claim, check if you include:
  \begin{enumerate}
    \item Statements of the full set of assumptions of all theoretical results. [Not Applicable]
    \item Complete proofs of all theoretical results. [Not Applicable]
    \item Clear explanations of any assumptions. [Not Applicable]
  \end{enumerate}

  \item For all figures and tables that present empirical results, check if you include:
  \begin{enumerate}
    \item The code, data, and instructions needed to reproduce the main experimental results (either in the supplemental material or as a URL). [Yes]
    \item All the training details (e.g., data splits, hyperparameters, how they were chosen). [Yes]
    \item A clear definition of the specific measure or statistics and error bars (e.g., with respect to the random seed after running experiments multiple times). [Yes]
    \item A description of the computing infrastructure used. (e.g., type of GPUs, internal cluster, or cloud provider). [Yes]
  \end{enumerate}

  \item If you are using existing assets (e.g., code, data, models) or curating/releasing new assets, check if you include:
  \begin{enumerate}
    \item Citations of the creator If your work uses existing assets. [Yes]
    \item The license information of the assets, if applicable. [Yes]
    \item New assets either in the supplemental material or as a URL, if applicable. [Yes]
    \item Information about consent from data providers/curators. [Not Applicable]
    \item Discussion of sensible content if applicable, e.g., personally identifiable information or offensive content. [Not Applicable]
  \end{enumerate}

  \item If you used crowdsourcing or conducted research with human subjects, check if you include:
  \begin{enumerate}
    \item The full text of instructions given to participants and screenshots. [Not Applicable]
    \item Descriptions of potential participant risks, with links to Institutional Review Board (IRB) approvals if applicable. [Not Applicable]
    \item The estimated hourly wage paid to participants and the total amount spent on participant compensation. [Not Applicable]
  \end{enumerate}

\end{enumerate}

\clearpage
\onecolumn
\appendix

\section*{Appendix}

\section{Model Details}

\subsection{Architecture} \label{appendix:model_architecture}

\begin{figure}
    \centering
    \includegraphics[width=0.75\textwidth]{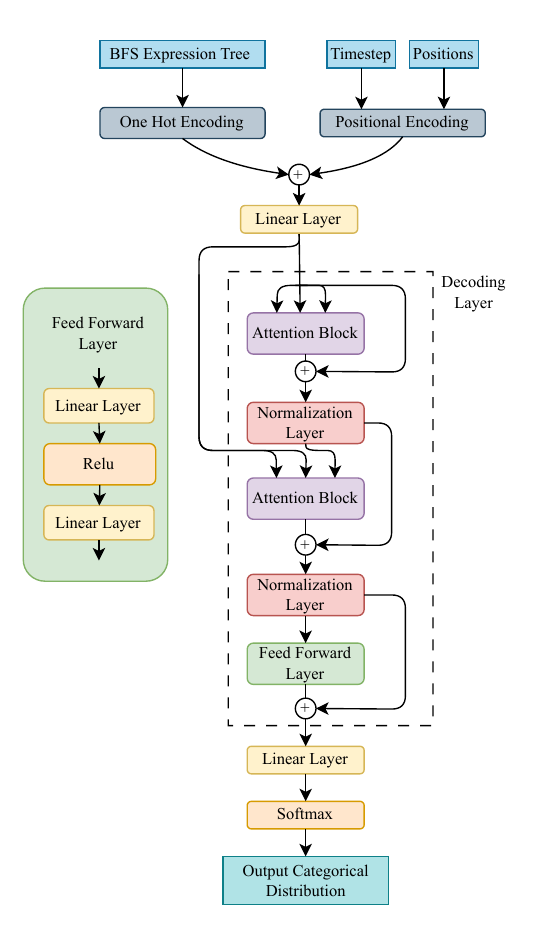}
    \caption{\ours's architecture.}
    \label{fig:mdsr_architecture}
\end{figure}

The full architecture of our model is shown in Figure \ref{fig:mdsr_architecture}. We leveraged a simple one-layer decoder only structure to focus our study to the sampling approach and policy optimization. The hyperparameter settings are listed in Table~\ref{tab:hyperparameters}. 


\begin{table} [!ht]
    \centering
    \caption{\small Hyperparameter settings of \ours.}
    \small
    \begin{tabular}{|c|c|}
    \hline
        \textit{Hyperparameter} & \textit{\ours} \\ \hline \hline
        Variables & \{1, $c$ (Constant Token), $xB_{i}$\} \\ \hline
        Unary Functions & \{sin, cos, log, $\sqrt{(\cdot)}$, exp\} \\ \hline
        Binary Functions & \{+, -, *, /, $\hat{}$ \} \\ \hline
        Risk Seeking Percent ($\alpha$) & $5\%$ \\ \hline
        Learning Rate & 5E-4  \\ \hline
        Max Depth & 32 \\ \hline
        Oversampling &  3  \\ \hline
        Entropy Coefficient $\lambda$ & 0.0005 \\ \hline
        Feed Forward Layers Size & 1024 \\ \hline
        Embedding Dim & 15 \\ \hline
        $C$ & 5 \\ \hline 
    \end{tabular}
    \label{tab:hyperparameters}
\end{table}

\subsection{Sampling Expressions}\label{sect:sampling-expr}

We sample an expression from a distribution matrix of size $M \times d$, where $M$ denotes the maximum length of the expression, and $d$ denotes the number of tokens in the library. The full sequence of $\tau$ is sampled before the expression tree is built. Once the sampling processes is complete, we remove any invalid tokens for $\tau$ and resample the invalid tokens from a uniform random sampling of valid tokens. This resampling avoids any of the generated expression from being invalid. Algorithm \ref{alg:mask_sampling} defines this process of sampling and generation expressions.
\ours further restricts that the sampled expression trees can include up to 10 constants. These constants are enforced in breadth-first search ordering. Additionally, nested $sin$ and $cos$ functions are prevented. Constant tokens are optimized with the Levenberg-Marquardt algorithm \citep{Levenberg_1944} for each discovered equation. The prevention of nested trigonometric functions is inherited from DSR, as the authors of DSR claimed nested trigonometric functions do not occur in physics. To approximate the DSR sampling procedure, we oversample by a factor of three during generation: for a batch size of $B$, we generate $3B$ candidate expressions (e.g., $3000$ for $B = 1000$) and then remove duplicates. If the number of unique expressions exceeds B, we truncate the sample to B. If fewer than B unique expressions remain, we retain duplicates to fill the batch. Crucially, we do not evaluate the oversampled expressions.

\begin{algorithm}[!h]
    \caption{\ours sampling of an expression}\label{alg:mask_sampling}
    \textbf{input} model $p_\theta$, sequence length $M$, vocabulary size $d$\\
    \textbf{output} An expression $\tau$
    \begin{algorithmic}[1]
        \STATE $X \gets \mathbf{0}^{M \times d}$ \hfill $\triangleright$ fully masked sequence (all-zero rows)
        \STATE $\mathcal{M} \gets \{0,\dots,M-1\}$ \hfill $\triangleright$ indices of masked positions
        \FOR{$t=M$ to $1$}
        \STATE $p \gets p_\theta(X, t)$ \hfill $\triangleright$ $p \in \mathbb{R}^{M \times d}$, one distribution per position
        \STATE $i \sim \mathcal{U}(\mathcal{M})$ \hfill $\triangleright$ pick a masked position uniformly at random
        \STATE $x^i \sim p_i$
        \STATE $X_i \gets$ one-hot encoding of $x^i$
        \STATE $\mathcal{M} \gets \mathcal{M} \setminus \{i\}$
        \ENDFOR
        \FOR{$i=0$ to $M-1$}
        \STATE $r \gets \text{Get Valid Tokens}(\tau, i)$
        \STATE $p_i \gets X_i \odot r$
        \IF{$\sum_{j=1}^{d} p_{i, j} = 0$}
        \STATE $p_i \gets r$
        \ENDIF
        \STATE $p_i \gets p_i / \sum_{j=1}^{d} p_{i, j}$
        \STATE $\tau_i \sim p_i$
        \ENDFOR
        \STATE \textbf{return} $\tau$
    \end{algorithmic}
\end{algorithm}

\begin{figure}
    \centering
    \includegraphics[width=0.35\textwidth]{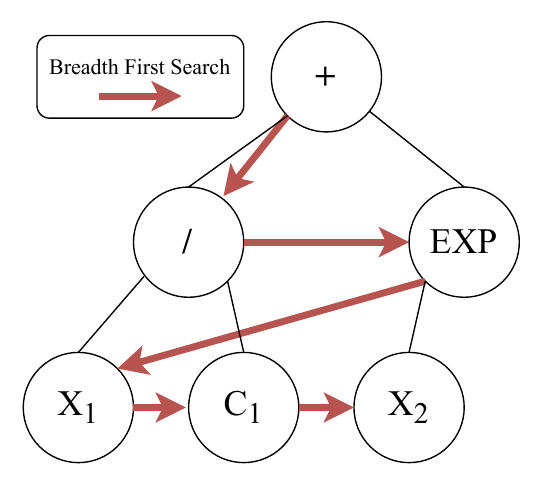}
    \caption{Breadth-first search ordering of the expression $\frac{x_1}{c_1} + \exp(x_2)$.}
    \label{fig:bfso_example}
\end{figure}

\textbf{Ordering: } Converting expression trees into a vector of tokens can be done in various ways. We used the tree's breadth-first search ordering (BFS) to order the nodes/tokens as done in CADSR. This ordering keeps siblings of parents nodes close to each other in the ordering. In comparison, the preorder traversal ordering (POT) can place siblings far apart. Figure \ref{fig:bfso_example} shows the conversion of an expression tree into a breadth-first search ordering and the comparison to a preorder traversal ordering.

\section{SRBench}

\subsection{Results}
Table \ref{tab:symbolic_problems_table} provides numerical results of SRBench for all of the methods we compared to in the paper and from the standard SRBench with bolded values denoting the best value in each column. EndToEnd, PySR, ParFam, and DL-ParFam are from \cite{scholl2025parfam} were they provided results for $0\%$ and $1\%$ noise levels. Additionally, we note that the reported time for EndToEnd does not align with the runtimes from \cite{kamienny_end--end_2022} and thus marked with an *. DL-ParFam is an alternative version of ParFam where the architecture of the network is generated by a foundation model, thus reducing runtime. 

\begin{table}[!h]
\centering
\caption{\small Performance on SRBench problems with known solutions.}
\label{tab:symbolic_problems_table}
\scriptsize
\begin{tabular}{lrrrrrrrrrrrrr}
\toprule
& \multicolumn{4}{c}{\textbf{Symbolic Solution Rate (\%)}} & \multicolumn{4}{c}{\textbf{Accuracy Rate (\%)}} & \multicolumn{4}{c}{\textbf{Simplified Complexity}} & \textbf{Time (m)}\\
\cmidrule(lr){2-5} \cmidrule(lr){6-9} \cmidrule(lr){10-13} \cmidrule{14-14}
\textbf{Algorithm} & 0.0 & 0.001 & 0.01 & 0.1 & 0.0 & 0.001 & 0.01 & 0.1 & 0.0 & 0.001 & 0.01 & 0.1 & Median \\
\midrule
AFP & 22.3 & 19.9 & 16.9 & 12.9 & 43 & 42 & 40 & 41 & 29.4 & 28.4 & 29.0 & 30.8 & 58.1 \\
AFP\_FE & 28.1 & 22.7 & 20.3 & 12.9 & 56 & 50 & 50 & 50 & 34.6 & 35.7 & 34.5 & 36.8 & 480.5 \\
AIFeynman & 61.8 & 31.9 & 12.6 & 0.9 & 74 & 74 & 68 & 10 & 83.3 & 88.7 & 99.3 & 110.5 & 493.2 \\
Bingo & 48.8 & 14.6 & 4.8 & 0.8 & 64 & 60 & 62 & 59 & 15.6 & 19.3 & 21.3 & 22.5 & 375.7 \\
BSR & 2.5 & 0.6 & 0.1 & 0.0 & 12 & 11 & 12 & 7 & 34.3 & 35.5 & 36.8 & 38.4 & 480.0 \\
CADSR& 41.8 & 34.3 & 33.6 & 29.2 & 55 & 54 & 55 & 54 & 16.3 & 15.7 & 15.9 & 15.0 & 360.0 \\
DL-ParFam & 45.9 & -- & 17.3 & -- & 84 & -- & 84 & -- & 185.8 & -- & 210.9 & -- & 1.77 \\
DSR & 24.8 & 24.4 & 17.5 & 10.5 & 38 & 41 & 40 & 39 & 16.6 & 16.1 & 17.0 & 18.5 & 452.2 \\
EndToEnd & 2.3 & -- & 0.0 & -- & 45 & -- & 36 & -- & 24.7 & -- & 28.3 & -- & 223* \\
EPLEX & 12.5 & 9.9 & 8.8 & 9.5 & 44 & 45 & 52 & 47 & 53.2 & 51.7 & 49.9 & 40.0 & 181.1 \\
FEAT & 0.1 & 0.0 & 0.0 & 0.0 & 40 & 43 & 41 & 14 & 88.0 & 77.3 & 72.6 & 50.4 & 18.0 \\
FFX & 0.0 & 0.0 & 0.0 & 0.1 & 0 & 0 & 3 & 18 & 274.9 & 273.3 & 286.0 & 341.4 & \textbf{0.3} \\
GP-GOMEA & 43.1 & 10.6 & 4.7 & 1.5 & 71 & 70 & 73 & 68 & 25.7 & 32.8 & 37.6 & 45.4 & 34.5 \\
ITEA & 20.8 & 13.8 & 7.7 & 1.5 & 27 & 27 & 27 & 26 & \textbf{14.5} & \textbf{15.0} & 15.4 & 16.0 & 23.5 \\
MDSR & \textbf{66.2} & \textbf{51.1} & 48.1 & \textbf{43.3} & 83 & 83 & 83 & \textbf{82} & 16.8 & 16.1 & 15.7 & \textbf{14.9} & 145.6 \\
MRGP & 0.0 & 0.0 & 0.0 & 0.0 & \textbf{93} & \textbf{92} & 89 & 2 & 110.0 & 106.5 & 83.1 & 0.0 & 308.8 \\
Operon & 16.0 & 12.3 & 1.9 & 0.1 & 87 & 86 & 86 & 73 & 40.8 & 40.1 & 60.4 & 70.8 & 41.4 \\
ParFam & 55.6 & -- & 41.4 & -- & \textbf{93} & -- & \textbf{90} & -- & 112.2 & -- & 143.9 & -- & 253.5 \\
PySR & 63.9 & -- & \textbf{53.0} & -- & 90 & -- & 87 & -- & 17.1 & -- & \textbf{13.9} & -- & 480.1 \\
SBP-GP & 22.7 & 0.7 & 0.0 & 0.0 & 74 & 74 & 75 & 54 & 109.9 & 102.1 & 112.9 & 116.3 & 482.8 \\
SNIP & -- & -- & -- & -- & -- & -- & -- & -- & 31.4 & 31.0 & 32.3 & 37.9 & 23.9 \\
TPSR & 36.1 & 0.0 & 0.0 & 0.0 & 68 & 68 & 66 & 38 & 57.1 & 59.3 & 63.4 & 65.8 & 2.9 \\
uDSR & 43.2 & -- & 18.4 & -- & 84 & -- & 79 & -- & 137.1 & -- & 167.9 & -- & 406.8 \\
gplearn & 16.2 & 16.9 & 16.6 & 16.0 & 30 & 29 & 27 & 22 & 45.8 & 37.8 & 36.4 & 33.8 & 23.3 \\
\bottomrule
\end{tabular}
\end{table}

\subsection{Noise}\label{sec:noise}

The noise levels  indicate the relative amount of noise added to the expression values for evaluation purposes. Specifically, for each data point, Gaussian noise is added as $y=f(x_1, \ldots, x_k) + \text{noise-level}\cdot s \cdot \epsilon$, where $f(x_1, \ldots, x_k)$ is the ground-truth expression, $s$ is the standard deviation of the expression values across all inputs $(x_1, \ldots, x_k)$ in the dataset, and $\epsilon$ is drawn from a standard Gaussian white noise distribution $\N(0, 1)$. This setup is the standard evaluation protocol in the SRBench benchmark, which is designed to assess the robustness of symbolic regression methods under varying noise levels.

\section{Real World Problems} \label{sec:realworld_appendix}

\subsection{Supplemental Hyperparameters} \label{sec:supplymental_hyperparameter}

We outline the hyperparameters and settings used for the methods evaluated on the real-world problems. Wherever possible, we followed the authors' recommended settings. \ours used the same hyperparameters as in our SRBench experiments. Explicit numerical results, including runtimes, are provided in Table \ref{tab:real-world-table-results}.

\textbf{CADSR.} We used the authors' suggested settings for real-world problems. CADSR used the same token library as \ours, and most of its remaining hyperparameters match those of \ours. It was evaluated with a batch size of 2000, 3 attention heads, and 2 decoder layers. CADSR also uses a Discrete Cosine Transform layer that reduces the embedding dimension from 15 to 12. For policy optimization, CADSR uses GRPO with 5 sampling rounds before each update of the reference policy $\theta_{\text{ref}}$, with $\beta=0.1$ and $\varepsilon=0.2$.

\textbf{PySR.} We ran PySR for 50{,}000 iterations with the same operator set as \ours, and used its default loss to select parsimonious expressions.

\textbf{SNIP.} SNIP's hyperparameters mainly control the search. We set the population size to 50, the maximum number of iterations to 80, and the beam size to 2.

\textbf{MODSR.} MODSR is a diffusion-based foundation model, and it is closely related to E2E. For generation, we used 50 diffusion steps and generated 20 candidate expressions.

\textbf{E2E.} E2E is a foundation model pretrained with a large token library, which is described in \citet{kamienny_end--end_2022}. We used a beam size of 10 and sampled 100 trees.

\begin{table*}[t]
    \centering
    \caption{Performance comparison on the water retention and woodrat datasets. We report the mean value across eight trials. Higher $R^2$ is better; lower runtime and complexity are better. Bolded entries denote best performance in the given column.}
    \label{tab:real-world-table-results}
    \small
    \begin{tabular}{llrrrr}
        \toprule
        Dataset & Method & Train $R^2$ & Test $R^2$ & Runtime (s) & Complexity \\
        \midrule
        \multirow{6}{*}{Water retention}
        & MODSR & 0.621 & 0.556 & 16.655    & 26.500 \\
        & E2E   & 0.000 & -0.042 & \textbf{8.168}     & 55.625  \\
        & SNIP  & 0.887 & 0.860 & 3558.539  & 35.250 \\
        & CADSR & 0.689 & 0.622 & 6481.074  & \textbf{17.625}\\
        & MDSR  & 0.864 & 0.835 & 1869.911  & 23.875 \\
        & PySR  & \textbf{0.967} & \textbf{0.948} & 3462.261  & 31.500\\
        \midrule
        \multirow{6}{*}{Woodrat}
        & MODSR & 0.573 & 0.421 & 11.0   & 34.9  \\
        & E2E   & 0.131 & 0.077 & \textbf{7.7} & 74.6 \\
        & SNIP  & 0.694 & 0.162 & 11756.1 & 63.4 \\
        & CADSR & 0.447 & 0.453 & 3514.7  & \textbf{5.2}  \\
        & MDSR  & 0.445 & \textbf{0.460} & 504.0   & 6.0  \\
        & PySR  & \textbf{0.778} & 0.219 & 3132.7  & 27.3 \\
        \bottomrule
    \end{tabular}

\end{table*}

\subsection{Pareto Frontier}

\begin{figure*}[t]
    \centering
    \begin{subfigure}[b]{0.49\textwidth}
        \centering
        \includegraphics[width=\textwidth]{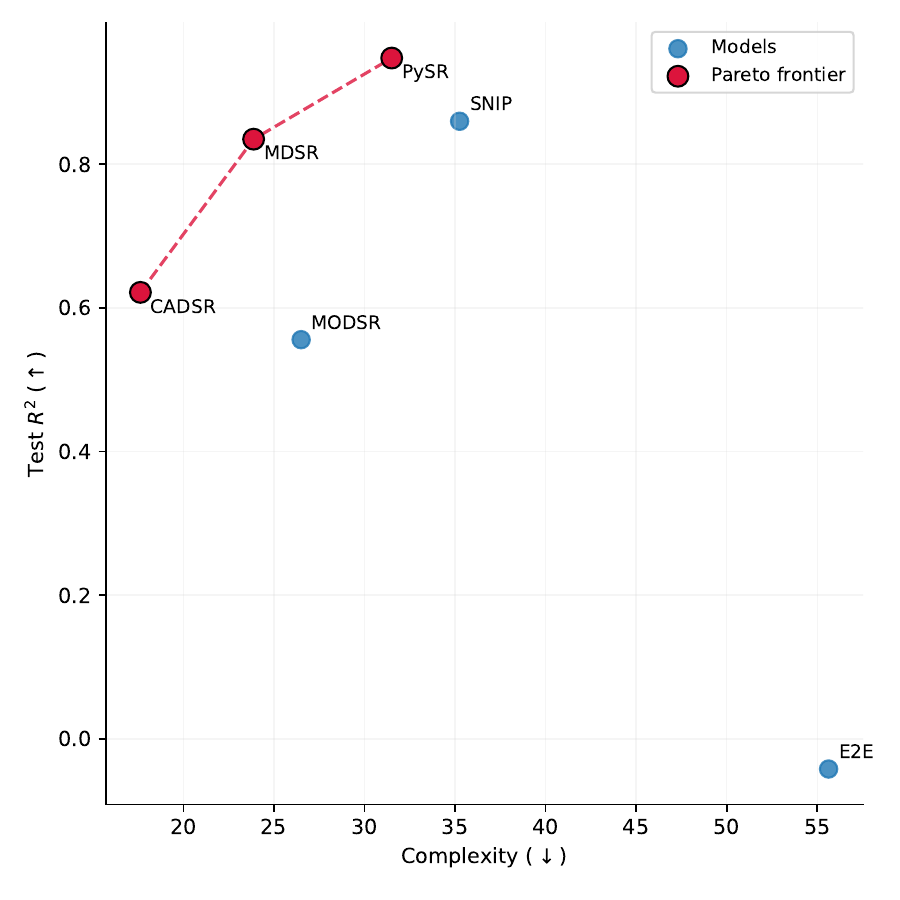}
        \caption{Mineral retention problem.}
        \label{fig:water-retention-pareto}
    \end{subfigure}
    \hfill
    \begin{subfigure}[b]{0.49\textwidth}
        \centering
        \includegraphics[width=\textwidth]{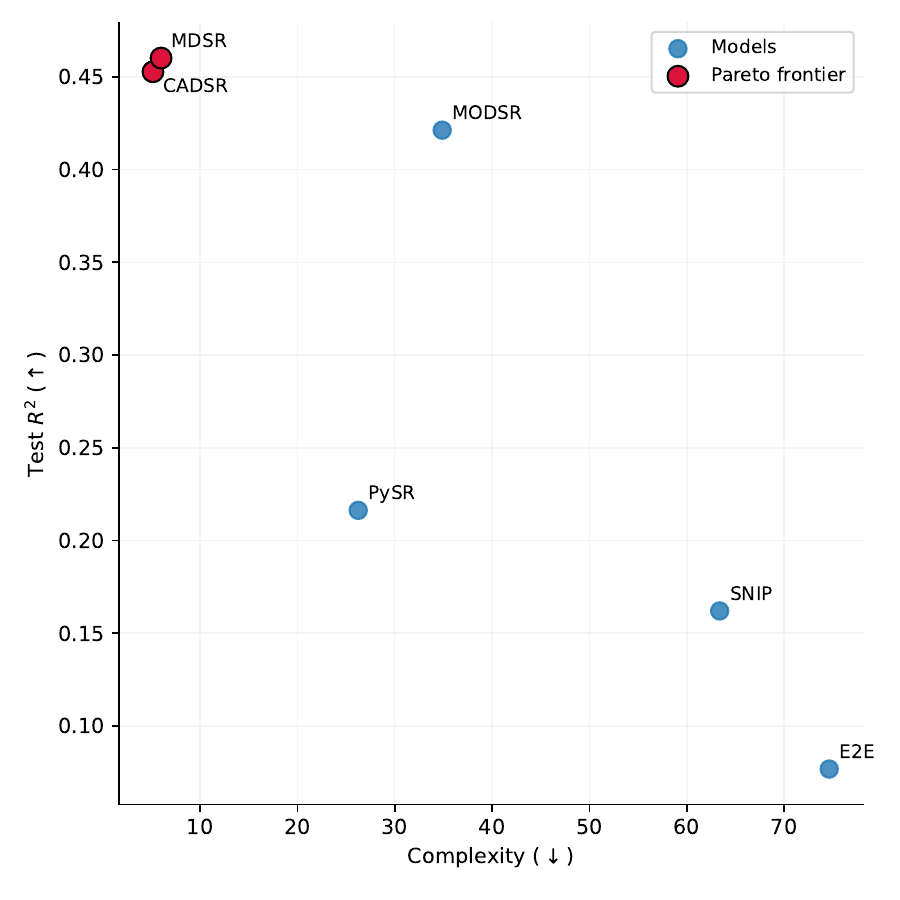}
        \caption{Genetic expression problem.}
        \label{fig:woodrat-pareto}
    \end{subfigure}
    \caption{Pareto-frontier comparison of symbolic-regression methods, showing test $R^2$ against expression complexity. Methods nearer the upper-left offer stronger predictive performance with simpler expressions.}
    \label{fig:pareto-frontiers}
\end{figure*}

The Pareto Frontier enable comparison of differing complexities to identify tradeoff between complexity and accuracy. Figures \ref{fig:water-retention-pareto} and \ref{fig:woodrat-pareto} compare $R^2$ test score with expression complexity. On both problems \ours shows up along the Pareto Frontier, while improving accuracy relative to CADSR. Additionally, while PySR appears on the Pareto Frontier for the mineral retention dataset it, overfits the gene expression dataset resulting in its fall off of the pareto frontier.

\subsection{Generated Expressions}

We provide the best expression found by each method in Table~\ref{tab:best_expressions_water_retention} and Table~\ref{tab:best_expressions_woodrat}. The results are selected by test $R^2$, so they are optimistic. For example, PySR's strong test score here may reflect selection bias: if enough expressions fit the training set, one of them is likely to fit the small test set well by chance. For the gene expression problem most methods fit the training data well, but many show a large gap between train and test accuracy. Together with the fact that much simpler expressions achieve comparable test accuracy, this gap suggests that the more complex expressions overfit the training set.

\newpage

\begin{table}[h]
\centering
\caption{Best expression found by each method on the mineral retention problem. Constants are rounded to three significant digits for readability.}
\label{tab:best_expressions_water_retention}
\scriptsize
\begin{tabular}{@{}l r c c p{0.55\linewidth}@{}}
\toprule
Model & Run & Train $R^2$ & Test $R^2$ & Best equation \\
\midrule
MODSR & 49529 & 0.724 & 0.792 &
\texttt{\seqsplit{(-1.35e4 + (((1.94 * sin((50.4 + (0.598 * PV)))) - (-0.351 * Flow)) - (((0.240 * Flow) - (-8.58e3 * arctan(((-5.43e3 * PV) - (741 * Size))))) - (-1.23 * Co))))}} \\
\addlinespace
E2E & 49529 & 0.004 & 0.117 &
\texttt{\seqsplit{(-0.005 + (-0.06 * ((1.6 + (-8.6 * sin(((0.01 + (0.2 * (1/(((-0.724 + (0.704 * (-1.26 + (0.378 * PV)))) * (-9 + (-0.09 * (-1.16 + (65.8 * Coductivity))))))))) + (8 + (-0.6 * cos(((0.09 + (0.9 * (-2.42 + (0.557 * Flow)))) + ((0.02 + (0.9 * (-17.3 + (16.7 * Size)))) + (0.02 + (9 * (-1.63 + (1.98 * Co))))))))))))**2))}} \\
\addlinespace
SNIP & 40320 & 0.972 & 0.975 &
\texttt{\seqsplit{(-6.16 + ((1/((-1.76e-4 + (-8.67e-5 * Co)))) * ((-2.28e-4 * Size) + ((1.47e-3 * Coductivity) - ((-2.15e-4 * arctan((-36.1 + (8.76 * PV)))) - ((-3.87e-4 * arctan((-9.90 + (12.5 * PV)))) - (3.23e-5 * Flow)))))))}} \\
\addlinespace
PySR & 49059 & 0.968 & 0.970 &
\texttt{\seqsplit{(1.09**(Flow/Co**0.853) + sin(1.83/(4.79 - PV)))*(sin(sin(2.36/PV)*1.40) + 0.602) - 4.99}} \\
\addlinespace
\ours & 8432 & 0.935 & 0.933 &
\texttt{\seqsplit{((c1 ** (PV + (((c3 + cos(PV)) + (log((Co / Flow)) + log(Co))) + (PV ** c2)))) - c0)}} \\
\addlinespace
CADSR & 8432 & 0.917 & 0.889 &
\texttt{\seqsplit{(((square(((c5 ** (PV ** PV)) / c4)) - (c3 / Size)) - (c1 / (PV ** (PV / (c6 ** Size))))) - ((c2 / PV) - c0))}} \\
\bottomrule
\end{tabular}
\end{table}

\begin{table}[h]
\centering
\caption{Best expression found by each method on the gene expression problem. Constants are rounded to three significant digits for readability.}
\label{tab:best_expressions_woodrat}
\scriptsize
\begin{tabular}{@{}l r c c p{0.55\linewidth}@{}}
\toprule
Model & Run & Train $R^2$ & Test $R^2$ & Best equation \\
\midrule
MODSR & 49529 & 0.680 & 0.532 &
\texttt{\seqsplit{(-5.28 + ((1.04 * PC2) - (-0.371 * (PC1 * (((-0.626 * Gsta1) - (0.0794 * Akr7a3)) - (((-8.59 * sin(((-0.207 * Gsta6) - (4.61 * cos(((6.56 * (Cyp3a11 * PC2)) - (-59.4 * Gsta3))))))) - (3.68 * Cyp2a3)) + (-0.911 * Cyp2a5)))))))}} \\
\addlinespace
E2E & 40320 & 0.144 & 0.341 &
\texttt{\seqsplit{((-5.2 + (0.36 * (0.002 + (9.42 * PC1)))) + ((0.004 + (-0.007 * cos(((-0.03 + (9 * (-3.68 + (0.902 * Cyp2a5)))) + (0.9 + (0.9 * (-4.09 + (0.332 * Cyp3a11)))))))) * ((0.01 + (60 * cos(((0.9 + (0.9 * (-5.62 + (0.598 * Akr7a3)))) * ((0.06 + (0.9 * (0.0476 + (9.52 * PC2)))) + ((0.01 + (-0.9 * (-2.4 + (0.92 * Gsta1)))) + (0.6 + (-2 * (-4.04 + (1.8 * Cyp2a3)))))))))) + ((-0.01 + (-0.01 * (-2.32 + (0.401 * Gsta6)))) + ((-0.01 + (-0.01 * (-2.31 + (0.924 * Gsta3)))) + (-0.5 + (0.01 * (-6.08 + (0.12 * Rsrc1)))))))))}} \\
\addlinespace
SNIP & 40320 & 0.687 & 0.504 &
\texttt{\seqsplit{(-5.23 + (((2.86 * PC2) - ((-0.271 * Cyp2a3) + ((9.56e-7 * (((-2.46e5 * Cyp2a3) + (9.98 * (Cyp3a11 * ((1.6e3 * Cyp2a3) - (87.5 * Akr7a3))))) * Gsta1)) - ((-34.3 * ((-0.0752 + (0.00184 * Cyp2a5)))**2) - (2.87e-9 * ((-1.23e4 + (-5.06e4 * PC2)))**2))))) * ((1/((-2.61e4 + (2.32e4 * Gsta1)))) * (((-0.751 * (Cyp3a11 * Cyp3a11)) + (5.68e3 * (1/((6.36e4 + (3.42e3 * Cyp3a11)))))) * ((412 * (Akr7a3 * PC1)) + (-6.49e3 * PC1))))))}} \\
\addlinespace
PySR & 49059 & 0.838 & 0.500 &
\texttt{\seqsplit{sin((PC1 + Gsta1**(-9.091))*(Cyp2a3*exp(sin(Gsta1**8.09) + cos(Rsrc1/0.595 + 0.011/PC1)) + sin(Gsta1**8.090))) - 5.229}} \\
\addlinespace
\ours & 49059 & 0.452 & 0.539 &
\texttt{\seqsplit{(c0 / (Cyp2a3 ** PC1))}} \\
\addlinespace
CADSR & 40320 & 0.475 & 0.541 &
\texttt{\seqsplit{((PC2 + c0) + tan(exp(PC1)))}} \\
\bottomrule
\end{tabular}
\end{table}

\newpage

\section{Ablations} \label{sec:ablations_appendix}
All ablations maintained similar hyperparameters to \ours in Table \ref{tab:hyperparameters} for direct comparisons.

\subsection{Policy Ablations} \label{sec:policy_ablations_appendix}

For PPO and GRPO we follow the hyperparameter settings for CADSR, leveraging 5 sampling steps, $G$, taken before updating the reference policy, $\theta_{\text{ref}}$, with $\beta=0.1$ and $\varepsilon = 0.2$. Algorithm \ref{alg:mdsr_with_grpo} provides the altered training algorithm for GRPO, and PPO is a drop in replacement for multi-step REINFORCE in Algorithm \ref{alg:ddsr}. All three methods REINFORCE, PPO, and GRPO are applied as a token-wise summation across tokens used in the expression $\tau^(i)$. For the one-step REINFORCE (denoted REINFORCE in ablations), we alter $C=1$ and use Algorithm \ref{alg:ddsr}.

\begin{align}
    \text{PPO}_\theta &=\frac{1}{B\alpha/100 }\sum\nolimits_{\tau^{(i)} \in \mathcal{S}_\alpha} \min\Big\{\frac{p_\theta(\tau^{(i)}|\X_t, t)}{p_{\theta_{\text{old}}}(\tau^{(i)}|\X_t, t)}, \text{clip}\big(\frac{p_\theta(\tau^{(i)}|\X_t, t)}{p_{\theta_{\text{old}}}(\tau^{(i)}|\X_t, t)}, 1-\epsilon, 1+\epsilon\big)\Big\} \cdot A_{i} \label{eq:PPO} \\
    \text{GRPO}_{\theta} &=\text{PPO}_\theta - \frac{1}{B\alpha/100 }\sum\nolimits_{\tau^{(i)} \in \mathcal{S}_\alpha}  \beta \text{KL}[p_\theta(\tau^{(i)}|\X_t, t) \| p_{\theta_{\text{ref}}}(\tau^{(i)}|\X_t, t)]  \label{eq:GRPO}
\end{align}

\begingroup
\setlength{\textfloatsep}{0pt}
\begin{algorithm}[!t]
	\caption{\ours with GRPO}
    \label{alg:mdsr_with_grpo}
	\textbf{input} Learning rate $\gamma$; risk factor $\alpha$; expression batch size $B$; steps per epoch $C$; epochs per reference $G$; number of epochs $N$; entropy scalar $\lambda$ \\
	\textbf{output} The best equation $\tau^*$ 
	\begin{algorithmic}[1]
		\STATE Initialize transformer $\phi$ with parameters $\theta$ \\
		\STATE $\Scal_\alpha \gets \{\}$ 
		\FOR{ $i =0$ to $N-1$}
		\IF{$ i\text{ mod}(G) = 0$}
		\STATE $\theta_{\text{ref}} \gets \theta$
		\ENDIF
		\STATE $\theta_{\text{old}} \gets \theta$
		\STATE Sample $B$ expression with the current model, and obtain the top $\alpha\%$ expressions $\Scal^i_\alpha$
		\STATE $\Scal_\alpha \gets \Scal_\alpha \cup \Scal^i_{\alpha}$
		\STATE Set $R_\alpha$ to the minimum award among the expressions in $\Scal_\alpha$. 
		\FOR{$j = 1$ to $C$}
		\STATE Randomly sample diffusion time step $t$ 
		\STATE Compute $J_{\text{GRPO}}$ from~\eqref{eq:GRPO}
        \STATE $\theta \gets \theta + \gamma (\nabla_{\theta} J_{\text{GRPO}} + \lambda\cdot \text{Entropy-Gradient})$
		\ENDFOR 
		\STATE Remove the bottom $\alpha\%$ expressions from $\Scal_\alpha$ according to the rewards
		\ENDFOR
		\STATE \textbf{return} the best expression from $\Scal_\alpha$
	\end{algorithmic}
\end{algorithm}
\endgroup

\begin{figure}[!h]
    \centering
    \includegraphics[width=0.5\textwidth]{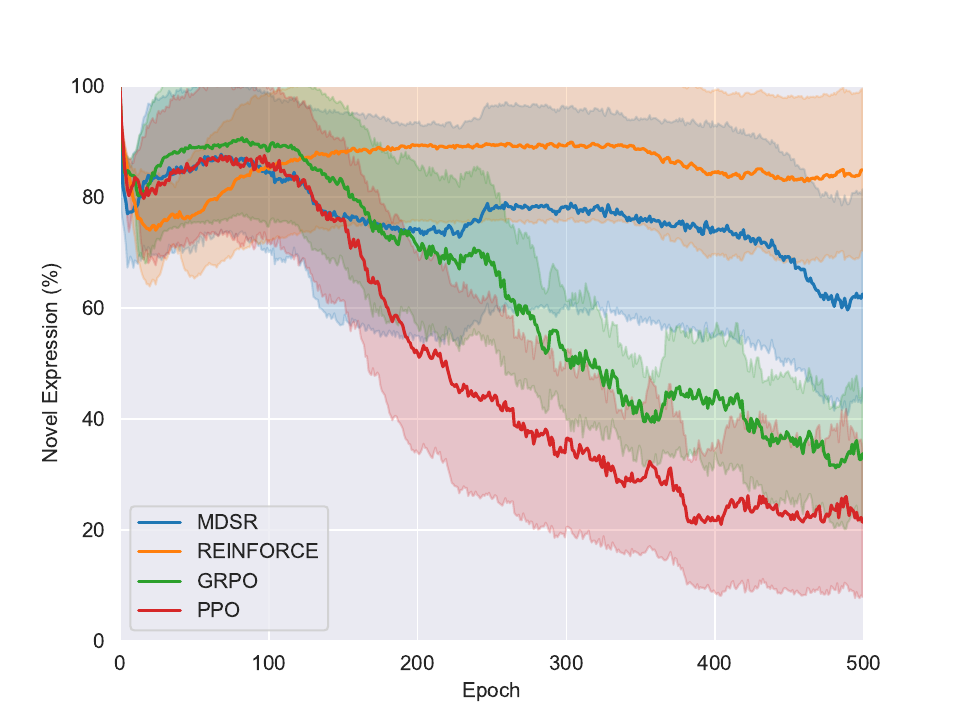}
    \caption{Impact of the policy optimization method on expression diversity. Multi-step REINFORCE, PPO, and GRPO all reduce diversity relative to single-step REINFORCE, with PPO and GRPO reducing diversity more than multi-step REINFORCE. While multi-step REINFORCE reduced diversity compared the single-step variant, it improve symbolic recovery which could highlight multi-steps REINFORCE's improvement to exploitation.}
    \label{fig:policy_diversity_ablation}
\end{figure}

\subsection{Sampling Ablations} \label{sec:sampling_ablations_appendix}
The sampling ablations focus on testing how deep symbolic regression functions based on sampling process. Below we provide an explicit implementation for each sampling method. All sampling methods are trained with the multi-step REINFORCE with the probability likelihood function tailor to the sampling method.

\textbf{Autoregressive.}
Autoregressive (AR) sampling is the standard approach for generating sequences: tokens are sampled in order of appearance, each conditioned on all previous tokens. A Beginning of Sequence (BOS) token is appended to the start of the sequence. Let $X = (BOS, x^1, \dots, x^{M-1})$ be a sequence of $M$ tokens. By the chain rule, the joint distribution factorizes as
\begin{equation}
    p_\theta(X) = \prod_{i=0}^{M-1} p_\theta\left(x^i \mid X^{0:i-1} \right), \label{eq:ar_factorization}
\end{equation}
so that token $i$ is sampled as $x^i \sim p_\theta(x^i \mid X^{0:i-1})$. Attention-based models enforce this factorization with a causal mask,
\begin{equation}
    A_{ij} =
    \begin{cases}
        0 & j \le i, \\
        -\infty & j > i,
    \end{cases} \label{eq:causal_mask}
\end{equation}
which is added to the attention logits so that position $i$ cannot attend to future positions. Because of this mask, all $M$ conditionals can be trained simultaneously in a single forward pass, with the loss in Equation \ref{eq:ar_loss} and the REINFORCE policy in Equation \ref{eq:ar_reinforce}.
\begin{align}
    \mathcal{L}_{\text{AR}} &= -\sum_{i=0}^{M-1} \log p_\theta\left(x^i \mid X^{0:i-1}\right) \label{eq:ar_loss}  \\
    \nabla_\theta J_{\text{MS}} &= -\frac{100}{B\alpha}\sum_{i=1}^{B} R(\tau^{(i)}) \nabla_\theta \log p_\theta\left(X^{1:M} \mid X^{0:M-1}\right) \label{eq:ar_reinforce}
\end{align}

AR sampling is a special case of our iterative masking framework (Equation~\ref{eq:masking_process}). In our method, a mask $m \in \{0,1\}^M$ is drawn from a masking distribution $Q$ and the model predicts the masked tokens given the unmasked ones. AR sampling corresponds to a deterministic $Q$ with a fixed left-to-right order: at step $i$, the mask is
\begin{equation}
    m^{(i)}_j = \mathbb{1}[j < i],
\end{equation}
so the first $i$ tokens are observed and the rest are masked, and the model predicts only token $i$. Our approach differs from AR sampling in three ways: (1) the order in which tokens are unmasked is random rather than fixed, (2) the causal attention mask is removed, so each prediction can condition on all currently unmasked tokens, including those to its right, and (3) the mask does not append the BOS token. 

For this ablation we also removed our 2D positional encoding (Equation~\ref{eq:positional_encoding}), which encodes both the token position $l$ and the masking iteration $t$, and replaced it with the standard 1D sinusoidal encoding of the token position alone,
\begin{align}
    \text{PE}(l)_{2i} &= \sin\left(\frac{l}{10000^{2i/D}}\right), \\
    \text{PE}(l)_{2i+1} &= \cos\left(\frac{l}{10000^{2i/D}}\right), \label{eq:positional_encoding_1d}
\end{align}
where $D$ is the embedding dimension. The iteration index $t$ is dropped because AR sampling has no masking iterations to distinguish. 

\begin{algorithm}[!h]
    \caption{Autoregressive sampling of an expression}\label{alg:ar_sampling}
    \textbf{input} model $p_\theta$, start token $x^0$, max length $M$\\
    \textbf{output} An expression $\tau$
    \begin{algorithmic}[1]
        \STATE $X^{0} \gets \text{BOS}$
        \FOR{$i=1$ to $M$}
        \STATE $p_i \gets p_\theta(\cdot \mid X^{0:i-1})$
        \STATE $r \gets \text{Get Valid Tokens}(\tau, i)$
        \STATE $p_i \gets p_i \odot r$
        \IF{$\sum_{j=1}^{d} p_{i, j} = 0$}
        \STATE $p_i \gets r$
        \ENDIF
        \STATE $p_i \gets p_i / \sum_{j=1}^{d} p_{i, j}$
        \STATE $x^i \sim p_i$
        \STATE $X^{0:i} \gets (X^{0:i-1}, x^i)$
        \ENDFOR
        \STATE $\tau \gets X^{1:M}$
        \STATE \textbf{return} $\tau$
    \end{algorithmic}
\end{algorithm}

\textbf{D3PM.}
The denoising diffusion probabilistic model (DDPM)~\citep{ho2020denoising} fundamentally shifted the paradigm of generative modeling, but it defines the forward process as a Gauss-Markov chain that gradually adds Gaussian noise and is therefore suited to continuous data. For categorical data such as discrete tokens, adding continuous Gaussian noise is neither feasible nor meaningful. To address this, \citet{austin2023d3pm} proposed the Discrete Denoising Diffusion Probabilistic Model (D3PM). Each data instance $\X_0 \in \mathbb{R}^{M \times d}$ represents a sequence of $M$ tokens, where each row is the one-hot encoding of a token over $d$ categories. D3PM defines a forward process that gradually transforms $\X_0$ toward a uniform distribution, effectively modeling discrete white noise. At each step $t>0$, the token distribution is updated as
\begin{equation}
    \X_t = \X_{t-1}\Q_t, \qquad \Q_t = \beta_t \I + (1-\beta_t)\frac{\mathbf{1}\mathbf{1}^\top}{d}, \label{eq:d3pm_forward}
\end{equation}
where $\mathbf{1}$ is a vector of ones and $\beta_t \in (0,1)$. Each row of $\X_t$ remains a valid probability distribution and converges to the uniform distribution as $t \to \infty$. The forward process has closed-form conditionals,
\begin{align}
    q(\X_t \mid \X_0) &= \X_0 \overline{\Q}_t, \qquad \overline{\Q}_t = \Q_1 \Q_2 \cdots \Q_t, \label{eq:d3pm_marginal} \\
    q(\X_{t-1} \mid \X_t, \X_0) &= \frac{\X_t \Q_t^\top \odot \X_0 \overline{\Q}_{t-1}}{\X_0 \overline{\Q}_t \X_t^\top}, \label{eq:d3pm_posterior}
\end{align}
where $\odot$ denotes element-wise multiplication.

During training, a timestep $t$ is drawn at random and corrupted tokens are sampled from $q(\X_t \mid \X_0)$. The corrupted tokens and $t$ are fed to a neural network $p_\theta(\X_0 \mid \X_t, t)$ that predicts the clean token distribution, trained with the cross-entropy loss in Equation \ref{eq:d3pm_loss}, and the REINFORCE policy in Equation \ref{eq:d3pm_reinforce}.
\begin{align}
    \mathcal{L}_{\text{D3PM}} &= -\mathbb{E}_{t,\, \X_t \sim q(\X_t \mid \X_0)} \left[ \log p_\theta(\X_0 \mid \X_t, t) \right]. \label{eq:d3pm_loss} \\
    \nabla_\theta J_{\text{MS}} &= -\frac{100}{B\alpha}\sum_{i=1}^{B} R(\tau^{(i)}) \nabla_\theta \log p_\theta(\X_0 \mid \X_t, t) \label{eq:d3pm_reinforce}
\end{align}
Generation begins with tokens sampled uniformly at random. At each step $t$, the reverse conditional is obtained by marginalizing $\X_0$ out of the posterior using the network's prediction,
\begin{equation}
    p_\theta(\X_{t-1} \mid \X_t) = \sum_{\X_0} q(\X_{t-1} \mid \X_t, \X_0)\, p_\theta(\X_0 \mid \X_t, t), \label{eq:d3pm_reverse}
\end{equation}
and $\X_{t-1}$ is drawn from it. This repeats until $t=0$, at which point $\X_0$ is the generated sample. Algorithm~\ref{alg:d3pm_sampling} outlines this procedure. In our ablation, we use $T=5$ diffusion steps, making D3PM the fastest sampling method, and use the same model architecture as \ours (including the 2D positional encoding in Equation \ref{eq:positional_encoding}).

D3PM is closely related to our iterative masking framework (Equation~\ref{eq:masking_process}). Both corrupt a sequence of tokens and train a network to recover the clean tokens from the corrupted ones. D3PM with an absorbing-state transition matrix, $\Q_t = \beta_t \I + (1-\beta_t)\mathbf{1}\mathbf{e}_{\texttt{[MASK]}}^\top$, reduces to a masking process in which each token is independently replaced by a mask token (nearly Mask-Diffusion). Our approach differs from the uniform-noise D3PM used here in two ways: (1) corrupted positions are marked with a mask token rather than replaced by random valid tokens, so the model knows which positions need to be predicted, (2) the corruption is uniform and aligns with the sequence length rather than controlled by the schedule $\{\beta_t\}$, and (3) the expression tree is built at each timestep.

\begin{algorithm}[!h]
    \caption{D3PM sampling of an expression}\label{alg:d3pm_sampling}
    \textbf{input} model $p_\theta$, number of steps $T$, sequence length $M$, vocabulary size $d$\\
    \textbf{output} An expression $\tau$
    \begin{algorithmic}[1]
        \STATE $\X_T \gets$ one-hot encoding of $M$ tokens drawn uniformly from $\{1,\dots,d\}$
        \FOR{$t=T$ to $1$}
            \STATE $\hat{\X}_0 \gets p_\theta(\X_0 \mid \X_t, t)$ \hfill $\triangleright$ predicted clean token distributions, $\hat{\X}_0 \in \mathbb{R}^{M \times d}$
            \FOR{$i=1$ to $M$}
                \STATE $p_{t-1}^{i} \gets \sum_{k=1}^{d} \hat{\X}_{0}^{i,k}\; q(x_{t-1}^{i} \mid x_t^{i}, x_0^{i} = e_k)$ \hfill $\triangleright$ Equation~\ref{eq:d3pm_reverse}
                \STATE $r \gets \text{Get Valid Tokens}(\tau, i)$
                \STATE $p_{t-1}^i \gets p_{t-1}^i \odot r$
                \IF{$\sum_{j=1}^{d} p_{t-1}^{i, j} = 0$}
                    \STATE $p_{t-1}^i \gets r$
                \ENDIF
                \STATE $p_{t-1}^i \gets p_{t-1}^i / \sum_{j=1}^{d} p_{t-1}^{i, j}$
                \STATE $x_{t-1}^{i} \sim p_{t-1}^{i}$
                \ENDFOR
            \STATE $\X_{t-1} \gets$ one-hot encoding of $(x_{t-1}^{1},\dots,x_{t-1}^{M})$
        \ENDFOR
        \STATE $\tau \gets \X_0$
        \STATE \textbf{return} $\tau$
    \end{algorithmic}
\end{algorithm}

\textbf{Mask-Diffusion.}
Mask-Diffusion combines the D3PM framework with masking through an absorbing-state transition, and is inspired by the masked-diffusion formulation of LLaDA~\citep{nie2025largelanguagediffusionmodels}. A masked token is represented by an all-zero row, and once a token is masked it cannot be unmasked in the forward process. At each forward step, a binary mask $Q_t \in \{0,1\}^{M}$ is drawn with independent entries,
\begin{equation}
    Q_t^i \sim \text{Bernoulli}(\beta_t), \qquad \beta_t = \frac{T-t}{T-t+1}, \label{eq:maskdiff_transition}
\end{equation}
where $Q_t^i = 1$ keeps token $i$ and $Q_t^i = 0$ masks it. The forward process is $X_t = Q_t \odot X_{t-1}$, applied row-wise, so each token is independently kept or masked. As in D3PM, the process has a closed-form marginal. A token survives to step $t$ with probability
\begin{equation}
    \bar{\alpha}_t = \prod_{s=1}^{t}\beta_s = \frac{T-t}{T}, \qquad q(X_t \mid X_0) = \bar{Q}_t \odot X_0, \quad \bar{Q}_t^i \sim \text{Bernoulli}(\bar{\alpha}_t). \label{eq:maskdiff_marginal}
\end{equation}
Consequently, the expected fraction of unmasked tokens at step $t$ is $(T-t)/T$, and the expected fraction of masked tokens is $t/T$.

In the reverse process, the network predicts the clean sequence from the current state, $\hat{X}_0 \sim p_\theta(X_0 \mid X_{t+1}, t+1)$, and a subset of the currently masked positions is revealed. Each masked token is revealed independently with probability $\frac{\bar{\alpha}_t - \bar{\alpha}_{t+1}}{1-\bar{\alpha}_{t+1}} = \frac{1}{t+1}$, while unmasked tokens remain fixed:
\begin{equation}
    X_t = X_{t+1} + R_t \odot \hat{X}_0, \label{eq:maskdiff_posterior}
\end{equation}
where $R_t$ is a Bernoulli reveal mask supported only on masked positions. Because the network predicts the full sequence at every step and the masking schedule controls which predictions are kept, the number of denoising steps is set by $T$. This gives the training loss
\begin{equation}
    \mathcal{L}_{\text{Mask-Diffu}} = -\mathbb{E}_{t,\, X_t \sim q(X_t \mid X_0)} \left[ \log p_\theta(X_0 \mid X_t, t) \right], \label{eq:maskdiff_loss}
\end{equation}
which we replace with the risk-seeking REINFORCE objective of Equation~\ref{eq:maskdiff_reinforce}, using $p_\theta(X_0 \mid X_t, t)$ as the policy.
\begin{align}
    \nabla_\theta J_{\text{MS}} &= -\frac{100}{B\alpha}\sum_{i=1}^{B} R(\tau^{(i)}) \nabla_\theta \log p_\theta(X_0 | X_t, t) \label{eq:maskdiff_reinforce}
\end{align}

Generation starts from a fully masked sequence, and at each step $t=T-1,\dots,0$ the network predicts all positions and a subset of the masked ones is revealed according to Equation~\ref{eq:maskdiff_posterior}. Algorithm \ref{alg:maskdiff_sampling} denotes this process. As in our other ablations, we use the same model architecture as \ours. We set the number of masking steps equal to the sequence length, $T=M$, so the expected number of unmasked tokens at step $t$ is $M-t$, matching that of our masking approach. However, the variance differs. In Mask-Diffusion each token is revealed independently, so the number of unmasked tokens at step $t$ is binomial and can be above or below $M-t$, whereas our masking approach enforces that the number of unmasked tokens equals the number of steps taken.

\begin{algorithm}[!h]
    \caption{Mask-Diffusion sampling of an expression}\label{alg:maskdiff_sampling}
    \textbf{input} model $p_\theta$, number of steps $T$, sequence length $M$, vocabulary size $d$\\
    \textbf{output} An expression $\tau$
    \begin{algorithmic}[1]
        \STATE $X_T \gets \mathbf{0}^{M \times d}$ \hfill $\triangleright$ fully masked sequence (all-zero rows)
        \FOR{$t=T-1$ to $0$}
        \STATE $p \gets p_\theta(X_0 \mid X_{t+1}, t+1)$ \hfill $\triangleright$ $p \in \mathbb{R}^{M \times d}$, one distribution per position
        \FOR{$i=1$ to $M$}
        \STATE $\hat{x}_0^{i} \sim p^{i}$
        \ENDFOR
        \STATE $\hat{X}_0 \gets$ one-hot encoding of $(\hat{x}_0^{1},\dots,\hat{x}_0^{M})$
        \FOR{$i=1$ to $M$}
        \IF{row $i$ of $X_{t+1}$ is masked}
        \STATE $Q_t^{i} \sim \text{Bernoulli}\left(\frac{1}{t+1}\right)$ \hfill $\triangleright$ Equation~\ref{eq:maskdiff_posterior}
        \ELSE
        \STATE $Q_t^{i} \gets 0$
        \ENDIF
        \ENDFOR
        \STATE $X_t \gets X_{t+1} + Q_t \odot \hat{X}_0$
        \ENDFOR
        \STATE $\tau \gets X_0$
        \STATE \textbf{return} $\tau$
    \end{algorithmic}
\end{algorithm}

\clearpage

\end{document}